# Planning over Chain Causal Graphs for Variables with Domains of Size 5 Is NP-Hard

**Omer Giménez**                                              OMER.GIMENEZ@UPC.EDU
*Dept. de Llenguatges i Sistemes Informàtics*
*Universitat Politècnica de Catalunya*
*Jordi Girona, 1-3*
*08034 Barcelona, Spain*

**Anders Jonsson**                                          ANDERS.JONSSON@UPF.EDU
*Dept. of Information and Communication Technologies*
*Universitat Pompeu Fabra*
*Roc Boronat, 138*
*08018 Barcelona, Spain*

## Abstract

Recently, considerable focus has been given to the problem of determining the boundary between tractable and intractable planning problems. In this paper, we study the complexity of planning in the class $\mathbb{C}_n$ of planning problems, characterized by unary operators and directed path causal graphs. Although this is one of the simplest forms of causal graphs a planning problem can have, we show that planning is intractable for $\mathbb{C}_n$ (unless P = NP), even if the domains of state variables have bounded size. In particular, we show that plan existence for $\mathbb{C}_n^k$ is NP-hard for $k \geq 5$ by reduction from CNF-SAT. Here, $k$ denotes the upper bound on the size of the state variable domains. Our result reduces the complexity gap for the class $\mathbb{C}_n^k$ to cases $k = 3$ and $k = 4$ only, since $\mathbb{C}_n^2$ is known to be tractable.

## 1. Introduction

There is an ongoing effort in the planning community to determine the complexity of different classes of planning problems. Known tractable classes are usually characterized by a simple causal graph structure accompanied by additional restrictions on variables and operators. However, the boundary between tractable and intractable planning problems is still not clearly established. The present paper contributes a novel complexity result for a class of planning problems with simple causal graph structure from the literature, in an effort to reduce this complexity gap.

The problem of determining tractable classes of planning problems is not purely of theoretical interest. For instance, complex planning problems can be projected onto tractable fragments of planning problems to generate heuristics to be used during search (Katz & Domshlak, 2008b). Also, the causal graph heuristic (Helmert, 2006) exploits the hierarchical structure of a planning problem by transforming it into a more tractable form: first, it translates propositional variables into multi-valued variables, a process that simplifies the causal graph of the problem; then, it keeps relaxing the problem until the causal graph becomes acyclic.

The present paper aims to study the complexity of planning problems in the class $\mathbb{C}_n$, defined by Domshlak and Dinitz (2001). The class $\mathbb{C}_n$ contains planning problems with multi-valued variables and chain causal graphs, i.e., the causal graph is just a directed path (implying that operators are unary). The notation $n$ indicates that the number of state variables is unbounded. In particular, we study the complexity of plan existence for $\mathbb{C}_n$, i.e., determining whether or not there exists a plan that solves a planning problem in $\mathbb{C}_n$.

| $\mathbb{C}_n^k$ | Plan generation | Macro plan generation | Plan existence |
|---|---|---|---|
| $k = 2$ | P | P | P |
| $k \in \{3, 4\}$ | EXP | ? | ? |
| $k \geq 5$ | EXP | **Intractable** | **NP-hard** |

Table 1: Overview of the complexity results for the class $\mathbb{C}_n^k$.

Even though planning problems in $\mathbb{C}_n$ exhibit an extremely basic form of causal structure, i.e., linear dependence between state variables, solving planning problems in $\mathbb{C}_n$ is not necessarily tractable, even if we impose additional restrictions. Let $\mathbb{C}_n^k$ be the subclass of $\mathbb{C}_n$ for which state variables have domains of size at most $k$. It is known that class $\mathbb{C}_n^2$ is polynomial-time solvable (Brafman & Domshlak, 2003) and that plan existence for class $\mathbb{C}_n$ is NP-hard (Giménez & Jonsson, 2008a). Our aim is to study the complexity of plan existence for those classes in between, namely $\mathbb{C}_n^k$ for $k \geq 3$.

Domshlak and Dinitz (2001) showed that there are solvable instances of $\mathbb{C}_n^3$ that require exponentially long plans. This means that there is no polynomial-time plan generation algorithm for $\mathbb{C}_n^k$ with $k \geq 3$, as was the case for $\mathbb{C}_n^2$. However, this does not rule out the existence of a polynomial-time algorithm that determines plan existence for class $\mathbb{C}_n^k$, or even an algorithm that generates plans in some succinct form, like those of Jonsson (2007) and Giménez and Jonsson (2008a). This is not incompatible with $\mathbb{C}_n$ being NP-hard.

In this paper, we prove that plan existence for the class $\mathbb{C}_n^k$ is NP-hard for $k \geq 5$. In other words, even if the causal graph is a directed path and the domains of the state variables are restricted to contain no more than 5 values, deciding whether or not a plan exists for solving the corresponding planning problem is NP-hard. Our result implies that it is not sufficient for a planning problem to exhibit linear variable dependence and restricted variable domain sizes; additional restrictions are necessary to make planning tractable.

Table 1 shows an overview of the complexity results for the class $\mathbb{C}_n^k$ to date. By "Macro plan generation" we mean any algorithm for generating a compact representation of the solution, such as in the work of Jonsson (2007) and Giménez and Jonsson (2008a). The "Intractable" result for this column means that the complexity is yet unknown but cannot be in P unless P = NP (else plan existence would be in P). The row for $k = 2$ is due to Brafman and Domshlak (2003), the column for plan generation is due to Domshlak and Dinitz (2001), and the contributions of the present paper are marked in boldface. Note that the novel result subsumes that of Giménez and Jonsson (2008a), who showed NP-hardness for $k = O(n)$.

This paper is organized as follows. In Section 2 we relate our results to previous work, and in Section 3 we introduce the notation used throughout. In Section 4 we give formal proof of a reduction from CNF-SAT to planning problems in $\mathbb{C}_n^{11}$. The main result, a reduction from CNF-SAT to planning problems in $\mathbb{C}_n^5$, is then proved in Section 5. Although the result for $\mathbb{C}_n^5$ subsumes that for $\mathbb{C}_n^{11}$, we believe that the intuitive idea behind the $\mathbb{C}_n^{11}$ reduction is easier to understand, and may be of interest for anyone trying to prove hardness results under similar circumstances. In Section 6 we discuss the complexity of the remaining classes $\mathbb{C}_n^3$ and $\mathbb{C}_n^4$.

We also prove the correctness of a third reduction, this time from CNF-SAT to $\mathbb{C}_n^7$, in Appendix A. The reductions for $\mathbb{C}_n^{11}$ and $\mathbb{C}_n^7$ previously appeared in a conference paper (Giménez & Jonsson, 2008b), and the present paper provides formal proof of their correctness.

## 2. Related Work

The complexity of planning has been studied extensively over the last twenty years (Bylander, 1994; Chapman, 1987; Erol, Nau, & Subrahmanian, 1995). Many tractable classes of planning problems exploit the notion of a causal graph in one way or another. Knoblock (1994) is usually credited with introducing the causal graph in his work on hierarchical planning. Williams and Nayak (1997) required planning problems to have acyclic causal graphs in an effort to ensure tractability. Jonsson and Bäckström (1998) defined the class 3S of planning problems, also with acyclic causal graphs, and showed that plan existence is tractable for this class.

Domshlak and Dinitz (2001) introduced the class $\mathbb{C}_n$ of planning problems studied in this paper, as well as several related classes, all of which have a particular causal graph structure. Brafman and Domshlak (2003) designed a polynomial-time algorithm for solving planning problems with binary state variables and polytree causal graphs of bounded indegree, proving that planning is tractable for the class $\mathbb{C}_n^2$. Brafman and Domshlak (2006) presented complexity results related to the tree-width of the causal graph. Katz and Domshlak (2008a) used causal graph structure to prove several complexity results for optimal planning.

Jonsson (2007) and Giménez and Jonsson (2008a) designed polynomial-time algorithms that solve planning problems with restricted causal graphs by generating a hierarchy of macros. Recently, Chen and Giménez (2008) showed that the complexity of planning is intractable unless the size of the largest connected component of the causal graph is bounded by a constant. Consequently, causal graph structure alone is not enough to guarantee tractability, implying that additional restrictions are needed.

## 3. Notation

Throughout the paper, we use $[i..n]$ to denote the set $\{i, \ldots, n\}$.

Let $V$ be a set of state variables, and let $D(v)$ be the finite domain of state variable $v \in V$. We define a state $s$ as a function on $V$ that maps each state variable $v \in V$ to a value $s(v) \in D(v)$ in its domain. A partial state $p$ is a function on a subset $V_p \subseteq V$ of state variables that maps each state variable $v \in V_p$ to $p(v) \in D(v)$. We frequently use the notation $(v_1 = x_1, \ldots, v_k = x_k)$ to denote a partial state $p$ defined by $V_p = \{v_1, \ldots, v_k\}$ and $p(v_i) = x_i$ for each $v_i \in V_p$.

A planning problem is a tuple $P = \langle V, \mathsf{init}, \mathsf{goal}, A \rangle$, where $V$ is the set of variables, $\mathsf{init}$ is an initial state, $\mathsf{goal}$ is a partial goal state, and $A$ is a set of operators. An operator $a = \langle pre(a); post(a) \rangle \in A$ consists of a partial state $pre(a)$ called the *pre-condition* and a partial state $post(a)$ called the *post-condition*. Operator $a$ is applicable in any state $s$ such that $s(v) = pre(a)(v)$ for each $v \in V_{pre(a)}$, and applying operator $a$ in state $s$ results in a new state $s'$ such that $s'(v) = post(a)(v)$ if $v \in V_{post(a)}$ and $s'(v) = s(v)$ otherwise.

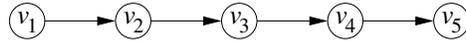

Figure 1: Example causal graph of a planning problem in the class $\mathbb{C}_5^k$.

A partial plan $\Pi$ for planning problem $P$ is a sequence of operators $a_1, \ldots, a_k \in A^k$, $k \geq 0$, such that $a_1$ is applicable in the initial state init and, for each $i \in [2..k]$, $a_i$ is applicable following the application of $a_1, \ldots, a_{i-1}$ starting in init. Note that a partial plan does not necessarily solve $P$. A plan $\Pi$ for solving $P$ is a partial plan such that the goal state goal is satisfied following the application of $a_1, \ldots, a_k$. $P$ is solvable if and only if there exists such a plan $\Pi$.

The causal graph of a planning problem $P$ is a directed graph $(V, E)$ with the state variables as nodes. There is an edge $(u, v) \in E$ if and only if $u \neq v$ and there exists an operator $a \in A$ such that $u \in V_{pre(a)} \cup V_{post(a)}$ and $v \in V_{post(a)}$. Figure 1 shows an example causal graph in the form of a directed path. The structure of the causal graph implies that each operator $a \in A$ is unary, i.e., the post-condition of $a$ is specified on a single variable $v$, and the pre-condition of $a$ is specified on (at most) $v$ and its predecessor $v'$ in the causal graph.

In this paper we study the class $\mathbb{C}_n^k$ of planning problems, defined as follows:

**Definition 3.1.** A planning problem $P$ belongs to the class $\mathbb{C}_n^k$ if and only if the causal graph of $P$ is a directed path and, for each $v \in V$, $|D(v)| \leq k$.

For planning problems in $\mathbb{C}_n^k$, the domain transition graph, or DTG, of a state variable $v$ is a labelled, directed graph $(D(v), E')$ with the values in the domain of $v$ as nodes. There is an edge $(x, y) \in E'$ with label $l \in D(v')$ if and only if there exists an operator $\langle v' = l, v = x; v = y \rangle$ in $A$, where $v'$ is the predecessor of $v$ in the causal graph. An edge without label indicates that the pre-condition of the corresponding operator is defined on $v$ alone. An edge with more than one label indicates the existence of multiple operators with the same pre- and post-condition on $v$ but different pre-conditions on $v'$.

## 4. $\mathbb{C}_n^{11}$ Is NP-hard

In this section we prove that $\mathbb{C}_n^{11}$ is NP-hard by reduction from CNF-SAT. In other words, to every CNF formula $F$ we associate a planning instance $P_{11}(F)$ of $\mathbb{C}_n^{11}$ such that $P_{11}(F)$ is solvable if and only if $F$ is satisfiable. We first describe the planning problem $P_{11}(F)$, then explain the intuitive idea behind the reduction, and finally provide formal proof of its correctness.

Let $F = C_1 \wedge \cdots \wedge C_k$ be a CNF formula on $k$ clauses and $n$ variables $x_1, \ldots, x_n$. We define the planning problem $P_{11}(F) = (V, \text{init}, \text{goal}, A)$ as follows. The variable set $V$ is $\{s_i \mid i \in [1..2n-1]\} \cup \{v_s\} \cup \{v_{ij} \mid i \in [1..k], j \in [1..n]\} \cup \{v_e\} \cup \{e_i \mid i \in [1..2n-1]\}$, with domains $D(s_i) = D(e_i) = D(v_e) = \{0, 1\}$ for $i \in [1..2n-1]$, $D(v_s) = \{0, 1, x\}$, and $D(v_{ij}) = \{g_x, g_0, g_1, a_x, a_0, a_1, b_0, b_1, c_x, c_0, c_1\}$ for $i \in [1..k], j \in [1..n]$. The initial state is defined by $\text{init}(s_i) = \text{init}(e_i) = \text{init}(v_e) = 0$, $i \in [1..2n-1]$, $\text{init}(v_s) = x$, and $\text{init}(v_{ij}) = a_x$ for $i \in [1..k], j \in [1..n]$, and the goal state is a partial state defined by $\mathsf{goal}(v_{in}) = g_x$ for each $i \in [1..k]$, $\mathsf{goal}(v_e) = 0$, and $\mathsf{goal}(e_i) = (i \bmod 2)$ for each $i \in [1..2n-1]$.

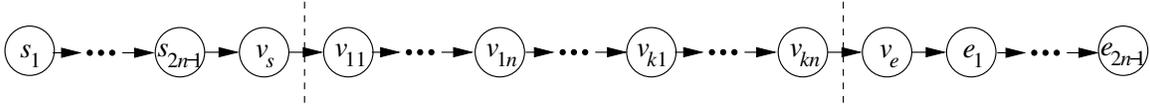

Figure 2: Causal graph of the planning problem $P_{11}(F)$.

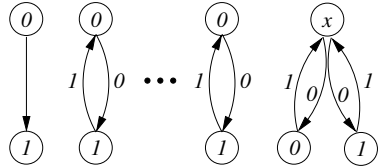

Figure 3: DTGs of the variables $s_1, s_2, \ldots, s_{2n-1}, v_s$.

Before providing a formal definition of the operators in $A$, we give an intuitive overview of the planning problem $P_{11}(F)$. To do this, we present the causal graph of $P_{11}(F)$ as well as the DTGs of each state variable. A reader who is only interested in the formal proof of the correctness of the reduction may skip to Section 4.2, where we introduce the formal definitions of operators in order to prove several theoretical properties of $P_{11}(F)$.

## 4.1 Intuition

The planning problem $P_{11}(F)$ associated to each CNF formula $F$ consists of three parts, each with a clearly defined role. The three parts are illustrated in Figure 2, showing the causal graph of $P_{11}(F)$. The first part of $P_{11}(F)$ corresponds to state variables $s_1, \ldots, s_{2n-1}, v_s$, the second part corresponds to state variables $v_{11}, \ldots, v_{1n}, \ldots, v_{k1}, \ldots, v_{kn}$, and the third part corresponds to state variables $v_e, e_1, \ldots, e_{2n-1}$. The role of the first part is to generate a message corresponding to an assignment to the variables of the CNF formula $F$. The role of the second part is to verify whether this assignment satisfies each clause $C_i$, and to remember this fact (using a value of state variable $v_{in}$). Finally, the role of the third part is to make sure that the message is propagated all the way to the end of the chain.

The DTGs of state variables $s_1, \ldots, s_{2n-1}, v_s$ appear in Figure 3. These state variables are used to generate an assignment $\sigma$ to the variables $x_1, \ldots, x_n$ of the CNF formula $F$. To do this, the operators of $P_{11}(F)$ are defined in such a way that the value of $v_s$ can change from $x$ to either 0 or 1, while from 0 or 1 it can only change back to $x$. Thus, by applying the operators of $P_{11}(F)$ it is possible to generate a sequence $x, m_1, x, \ldots, x, m_n, x$ of values of $v_s$, where $m_j \in \{0, 1\}$ for each $j \in [1..n]$.

We define a message $m$ as the sequence $m_1, \ldots, m_n$ of $n$ symbols (either 0 or 1) corresponding to a sequence of values of $v_s$. In what follows, we refer to these symbols as the "bits" of the message. The value $x$ is used as a separator to distinguish between consecutive bits of the message. Given a message $m$, the assignment $\sigma$ is defined as $\sigma(x_j) = m_j$ for each $j \in [1..n]$. Thus, the assignment to $x_1$ is determined by the first choice of whether to change the value of $v_s$ from $x$ to 0 or 1, and so on. The only purpose of the remaining state variables $s_i$ of the first part is to restrict the message $m$ to contain no more than $n$ bits.

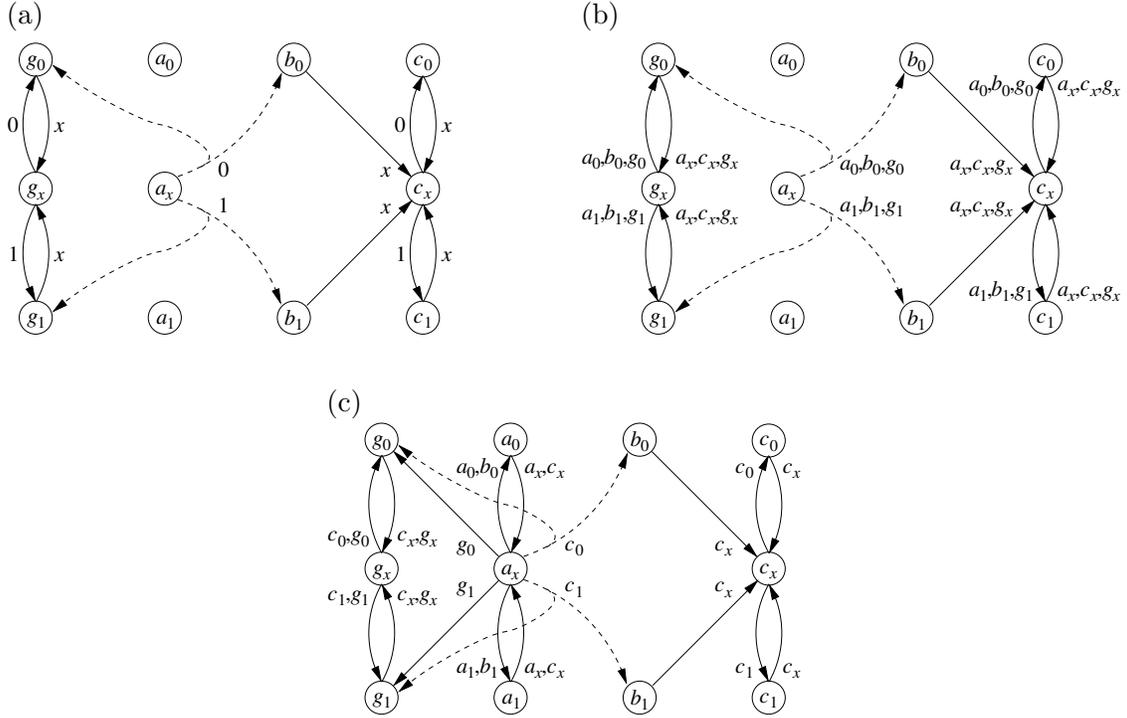

Figure 4: DTGs of (a) $v_{11}$, (b) $v_{i1}$ for $i > 1$, and (c) $v_{ij}$ for $j > 1$. Dashed edges are explained in the text.

The DTGs of state variables $v_{ij}$, $i \in [1..k]$ and $j \in [1..n]$, appear in Figure 4. The dashed edges in the DTGs indicate that the corresponding operators depend on the CNF formula $F$. For example, if the assignment $\sigma(x_1) = 1$ satisfies the clause $C_1$, the edge from $v_{11} = a_x$ with label 1 in Figure 4(a) points to $g_1$, else it points to $b_1$. Likewise, if $\sigma(x_1) = 0$ satisfies $C_1$, the edge from $v_{11} = a_x$ with label 0 points to $g_0$, else it points to $b_0$.

Recall that the role of the second part is to check whether the assignment $\sigma$ generated by the first part satisfies the CNF formula $F$. For each clause $C_i$ and each variable $x_j$ of $F$, the main function of state variable $v_{ij}$ is to check whether the assignment $\sigma(x_j) = m_j$ satisfies $C_i$. To do this, state variable $v_{ij}$ acts as a finite state automaton that propagates each bit of the message $m$ while keeping track of when the $j$-th bit of the message arrives. Since the domain size of state variables is restricted, there is no way for $v_{ij}$ to count the number of bits it has received. Instead, the fact that the $j$-th bit has arrived is indicated to it by $v_{i(j-1)}$. Moreover, the last state variable $v_{in}$ for each clause $C_i$ has to remember whether or not $C_i$ has been satisfied by the assignment to some variable $x_j$.

In summary, each state variable $v_{ij}$ in the second part performs the following functions through its values and operators:

1. Propagate the message $m$ generated by $v_s$.

2. Check whether the assignment to $x_j$ (the $j$-th bit of $m$) satisfies the clause $C_i$.

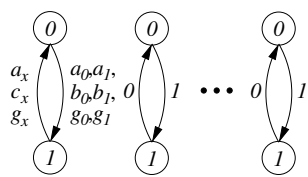

Figure 5: The domain transition graph of the variables $v_e, e_1, \ldots, e_{2n-1}$.

3. Remember whether $C_i$ was satisfied by the assignment to some $x_l$, $l \leq j$.

4. If $j < n$ and $C_i$ has been satisfied, propagate this fact.

5. If $j < n$, let $v_{i(j+1)}$ know when the $(j+1)$-th bit of the message has arrived.

Note that the third function is only strictly necessary for $j = n$. However, including it for all state variables makes the reduction more compact because of symmetry.

Next, we briefly describe how $v_{ij}$ implements each of these functions. Each value in the domain of $v_{ij}$ has subscript 0, 1, or $x$. To propagate the message, $v_{ij}$ always moves to a value whose subscript matches that of its predecessor (in the case of $v_{11}$, its subscript should match the value of $v_s$). Unless $C_i$ is satisfied by the assignment to $x_l$, $l < j$, the value of $v_{ij}$ remains in the subdomain $\{a_0, a_1, a_x\}$ prior to the arrival of the $j$-th bit.

The clause $C_i$ is encoded into the dashed edges of the DTGs of variables $v_{ij}$. These operators are such that when the $j$-th bit $m_j$ arrives, $v_{ij}$ moves from $a_x$ to $g_{m_j}$ if the assignment $\sigma(x_j) = m_j$ satisfies $C_i$, and to $b_{m_j}$ otherwise. The fact that the value of $v_{ij}$ is in the subdomain $\{g_0, g_1, g_x\}$ indicates that $C_i$ was satisfied by the assignment to some $x_l$, $l \leq j$. This fact is propagated all the way to $v_{in}$ since each subsequent state variable for $C_i$ is forced to move to a value in the subdomain $\{g_0, g_1, g_x\}$ whenever the value of its predecessor is in $\{g_0, g_1, g_x\}$. Whether or not a clause $C_i$ has been satisfied is checked by defining a goal state $v_{in} = g_x$.

Finally, if $j < n$ and $v_{ij}$ moves to $b_{m_j}$, then $v_{i(j+1)}$ moves to $a_{m_j}$. From there, $v_{ij}$ has no choice but to move to $c_x$, causing $v_{i(j+1)}$ to return to $a_x$. When the next bit arrives, $v_{ij}$ moves to either $c_0$ or $c_1$, correctly indicating to $v_{i(j+1)}$ that the $(j+1)$-th bit has arrived. Consequently, $v_{i(j+1)}$ moves to either $g_0$ ($g_1$) or $b_0$ ($b_1$), depending on whether or not the assignment to $x_{j+1}$ satisfies $C_i$. Hence, the values of type $b$ are used to *delay* the transition of $v_{i(j+1)}$ from a value of type $a$ to either $b$ or $g$. This is the mechanism that allows a variable $v_{ij}$ to react to the $j$-th bit. For each clause $C_i$, the operators for $v_{i1}$ are defined such that $v_{i1}$ always reacts to the first bit.

The DTGs of state variables $v_e, e_1, \ldots, e_{2n-1}$ appear in Figure 5. The function of these state variables is to make sure that all $n$ bits of the message $m$ are propagated to the end of the causal graph. A state variable (strictly speaking, a planner solving the planning problem) is never forced to select an operator, so it can choose not to propagate a bit of the message and instead wait for the next bit to arrive before acting. In turn, this may cause another state variable to incorrectly conclude that a clause has (not) been satisfied. The variables of the third part prevent this from happening, since the goal state is defined in such a way that it cannot be reached unless all bits of the message arrive at the end of the causal graph.

| Variable | Operator | Qualifier |
|---|---|---|
| $s_1$ | $\langle s_1 = 0; s_1 = 1 \rangle$ | |
| $s_i$, | $\langle s_{i-1} = 0, s_i = 0; s_i = 1 \rangle$ | |
| $i \in [2..2n-1]$ | $\langle s_{i-1} = 1, s_i = 1; s_i = 0 \rangle$ | |
| $v_s$ | $\langle s_{2n-1} = 0, v_s = x; v_s = m \rangle$ | $m \in \{0,1\}$ |
| | $\langle s_{2n-1} = 1, v_s = m; v_s = x \rangle$ | $m \in \{0,1\}$ |

Table 2: Operators for the variables $s_1, s_2, \ldots, s_{2n-1}, v_s$.

## 4.2 Formal Proof

In this section, we prove that $\mathbb{C}_n^{11}$ is NP-hard by showing that the planning problem $P_{11}(F)$ is solvable if and only if the formula $F$ has a satisfying assignment. To start with, we provide formal definitions of the operators of $P_{11}(F)$. The operators for $s_1, \ldots, s_{2n-1}, v_s$ appear in Table 2, and the corresponding DTGs appear in Figure 3. The operators for variables $v_{ij}$, $i \in [1..k]$ and $j \in [1..n]$, appear in Table 3, and the DTGs appear in Figure 4. Finally, the operators for $v_e, e_1, \ldots, e_{2n-1}$ appear in Table 4, and the DTGs appear in Figure 5.

To reduce the space requirement we use shorthand in the definitions of operators. In other words, $\langle v' = m, v = c; v = m \rangle$, $m \in \{a, b\}$, denotes the existence of two operators $\langle v' = a, v = c; v = a \rangle$ and $\langle v' = b, v = c; v = b \rangle$. Similarly, $\langle v' \in \{a, b\}, v = c; v = d \rangle$ denotes the existence of two operators $\langle v' = a, v = c; v = d \rangle$ and $\langle v' = b, v = c; v = d \rangle$. For state variables $v_{ij}$ we also introduce reference numbers that allow us to easily refer to operators.

Furthermore, some operators are conditional on properties of the CNF formula $F$; such an operator only exists if the indicated property is satisfied. For example, the operator $\langle v_{22} = c_0, v_{23} = a_x; v_{23} = g_0 \rangle$ only exists if the clause $C_2$ is satisfied by $\overline{x}_3$, and the operator $\langle v_{22} = c_0, v_{23} = a_x; v_{23} = b_0 \rangle$ only exists if $C_2$ is not satisfied by $\overline{x}_3$. We use the set notation $x_j \in C_i$ to denote that the literal $x_j$ appears in the clause $C_i$.

The proof is organized as follows. We begin with a series of technical definitions and lemmas (4.1–4.6) related to the operators and their implications. Definition 4.7 then introduces the notion of *admissible* plans, and Lemma 4.8 states that any plan for solving $P_{11}(F)$ has to be admissible. Next, Lemma 4.10 establishes that any admissible plan corresponds to an assignment to the variables of the CNF formula $F$, and that all operator choices of the plan are forced given this assignment. Finally, Lemma 4.13 determines the exact sequence of values taken on by each state variable during the execution of an admissible plan, making it possible to check whether the goal state is reached at the end of the execution. Theorem 4.14 then concludes that the only admissible plans solving $P_{11}(F)$ are those corresponding to satisfying assignments of $F$.

**Definition 4.1.** Given a partial plan $\Pi$ for $P_{11}(F)$ and a variable $v \in V$, $\Pi(v)$ is the number of times the value of $v$ is changed by operators in $\Pi$.

**Lemma 4.2.** *For each partial plan $\Pi$ for $P_{11}(F)$, it holds that*

- $\Pi(s_i) \leq i$ *for $i \in [1..2n-1]$, and*

- $\Pi(v_s) \leq 2n$.

| Variable | Ref. | Operator | Qualifier |
|---|---|---|---|
| $v_{11}$ | (1) | $\langle v_s = 1, v_{11} = a_x; v_{11} = g_1 \rangle$ | $x_1 \in C_1$ |
| | (2) | $\langle v_s = 1, v_{11} = a_x; v_{11} = b_1 \rangle$ | $x_1 \notin C_1$ |
| | (3) | $\langle v_s = 0, v_{11} = a_x; v_{11} = g_0 \rangle$ | $\overline{x}_1 \in C_1$ |
| | (4) | $\langle v_s = 0, v_{11} = a_x; v_{11} = b_0 \rangle$ | $\overline{x}_1 \notin C_1$ |
| | (5) | $\langle v_s = m, v_{11} = c_x; v_{11} = c_m \rangle$ | $m \in \{0, 1\}$ |
| | (6) | $\langle v_s = m, v_{11} = g_x; v_{11} = g_m \rangle$ | $m \in \{0, 1\}$ |
| | (7) | $\langle v_s = x, v_{11} = b_m; v_{11} = c_x \rangle$ | $m \in \{0, 1\}$ |
| | (8) | $\langle v_s = x, v_{11} = c_m; v_{11} = c_x \rangle$ | $m \in \{0, 1\}$ |
| | (9) | $\langle v_s = x, v_{11} = g_m; v_{11} = g_x \rangle$ | $m \in \{0, 1\}$ |
| $v_{i1}$, | (1) | $\langle v_{(i-1)n} \in \{a_1, b_1, g_1\}, v_{i1} = a_x; v_{i1} = g_1 \rangle$ | $x_1 \in C_i$ |
| $i \in [2..k]$ | (2) | $\langle v_{(i-1)n} \in \{a_1, b_1, g_1\}, v_{i1} = a_x; v_{i1} = b_1 \rangle$ | $x_1 \notin C_i$ |
| | (3) | $\langle v_{(i-1)n} \in \{a_0, b_0, g_0\}, v_{i1} = a_x; v_{i1} = g_0 \rangle$ | $\overline{x}_1 \in C_i$ |
| | (4) | $\langle v_{(i-1)n} \in \{a_0, b_0, g_0\}, v_{i1} = a_x; v_{i1} = b_0 \rangle$ | $\overline{x}_1 \notin C_i$ |
| | (5) | $\langle v_{(i-1)n} \in \{a_m, b_m, g_m\}, v_{i1} = c_x; v_{i1} = c_m \rangle$ | $m \in \{0, 1\}$ |
| | (6) | $\langle v_{(i-1)n} \in \{a_m, b_m, g_m\}, v_{i1} = g_x; v_{i1} = g_m \rangle$ | $m \in \{0, 1\}$ |
| | (7) | $\langle v_{(i-1)n} \in \{a_x, c_x, g_x\}, v_{i1} = b_m; v_{i1} = c_x \rangle$ | $m \in \{0, 1\}$ |
| | (8) | $\langle v_{(i-1)n} \in \{a_x, c_x, g_x\}, v_{i1} = c_m; v_{i1} = c_x \rangle$ | $m \in \{0, 1\}$ |
| | (9) | $\langle v_{(i-1)n} \in \{a_x, c_x, g_x\}, v_{i1} = g_m; v_{i1} = g_x \rangle$ | $m \in \{0, 1\}$ |
| $v_{ij}$, | (10) | $\langle v_{i(j-1)} = c_1, v_{ij} = a_x; v_{ij} = g_1 \rangle$ | $x_j \in C_i$ |
| $i \in [1..k]$, | (11) | $\langle v_{i(j-1)} = c_1, v_{ij} = a_x; v_{ij} = b_1 \rangle$ | $x_j \notin C_i$ |
| $j \in [2..n]$ | (12) | $\langle v_{i(j-1)} = c_0, v_{ij} = a_x; v_{ij} = g_0 \rangle$ | $\overline{x}_j \in C_i$ |
| | (13) | $\langle v_{i(j-1)} = c_0, v_{ij} = a_x; v_{ij} = b_0 \rangle$ | $\overline{x}_j \notin C_i$ |
| | (14) | $\langle v_{i(j-1)} \in \{a_m, b_m\}, v_{ij} = a_x; v_{ij} = a_m \rangle$ | $m \in \{0, 1\}$ |
| | (15) | $\langle v_{i(j-1)} = g_m, v_{ij} = a_x; v_{ij} = g_m \rangle$ | $m \in \{0, 1\}$ |
| | (16) | $\langle v_{i(j-1)} = c_m, v_{ij} = c_x; v_{ij} = c_m \rangle$ | $m \in \{0, 1\}$ |
| | (17) | $\langle v_{i(j-1)} \in \{c_m, g_m\}, v_{ij} = g_x; v_{ij} = g_m \rangle$ | $m \in \{0, 1\}$ |
| | (18) | $\langle v_{i(j-1)} \in \{a_x, c_x\}, v_{ij} = a_m; v_{ij} = a_x \rangle$ | $m \in \{0, 1\}$ |
| | (19) | $\langle v_{i(j-1)} = c_x, v_{ij} = b_m; v_{ij} = c_x \rangle$ | $m \in \{0, 1\}$ |
| | (20) | $\langle v_{i(j-1)} = c_x, v_{ij} = c_m; v_{ij} = c_x \rangle$ | $m \in \{0, 1\}$ |
| | (21) | $\langle v_{i(j-1)} \in \{c_x, g_x\}, v_{ij} = g_m; v_{ij} = g_x \rangle$ | $m \in \{0, 1\}$ |

Table 3: Operators for the variables $v_{11}, \ldots, v_{kn}$.

| Variable | Operator |
|---|---|
| $v_e$ | $\langle v_{kn} \in \{a_0, a_1, b_0, b_1, g_0, g_1\}, v_e = 0; v_e = 1 \rangle$ |
| | $\langle v_{kn} \in \{a_x, c_x, g_x\}, v_e = 1; v_e = 0 \rangle$ |
| $e_1$ | $\langle v_e = 1, e_1 = 0; e_1 = 1 \rangle$ |
| | $\langle v_e = 0, e_1 = 1; e_1 = 0 \rangle$ |
| $e_i, i \in [2..2n-1]$ | $\langle e_{i-1} = 1, e_i = 0; e_i = 1 \rangle$ |
| | $\langle e_{i-1} = 0, e_i = 1; e_i = 0 \rangle$ |

Table 4: Operators for the variables $v_e, e_1, \ldots, e_{2n-1}$.

*Proof.* By induction on $i$. For $i = 1$, variable $s_1$ can only change once, so $\Pi(s_1) \leq 1$. For $i \in [2..2n-1]$, it follows from inspection of the operators that we cannot change the value of $s_i$ twice without changing the value of $s_{i-1}$ once in between (the operator for setting $s_i$ to 1 has $s_{i-1} = 0$ as a pre-condition, and the operator for resetting $s_i$ to 0 has $s_{i-1} = 1$ as a pre-condition). Since we can change the value of $s_i$ once in the initial state without first changing the value of $s_{i-1}$, it follows that $\Pi(s_i) \leq \Pi(s_{i-1}) + 1 \leq (i-1) + 1 = i$ by induction. The same argument holds for variable $v_s$ and its predecessor $s_{2n-1}$, so $\Pi(v_s) \leq \Pi(s_{2n-1}) + 1 \leq (2n-1) + 1 = 2n$. ∎

**Lemma 4.3.** *For each partial plan $\Pi$ for $P_{11}(F)$ and each $v_{ij}$, $i \in [1..k]$ and $j \in [1..n]$, it holds that $\Pi(v_{ij}) \leq \Pi(v')$, where $v'$ is the predecessor of $v_{ij}$ in the causal graph.*

*Proof.* Just as before, it follows by inspection of the operators that we cannot change the value of $v_{ij}$ twice without changing the value of $v'$ in between. To see this, note that the subscript of each value in $D(v_{ij})$ is either $x$, 0, or 1. An operator for $v_{ij}$ either changes its value from one with subscript $x$ to one with subscript 0 (1), if $v'$ also has a value with subscript 0 (1), or from one with subscript 0 (1) to one with subscript $x$, if $v'$ also has a value with subscript $x$ (the same argument holds for $v_{11}$, although the values of its predecessor $v_s$ are $x$, 0, and 1 without subscripts).

Note that the value of $v_{ij}$ cannot change in the initial state without first changing the value of $v'$, since $v'$ has to have a value with subscript 0 or 1 for the value of $v_{ij}$ to change from its initial value $a_x$. Consequently, the value of $v_{ij}$ cannot change more times than the value of $v'$, so $\Pi(v_{ij}) \leq \Pi(v')$ as claimed. ∎

**Lemma 4.4.** *For each $v_{ij}$, $i \in [1..k]$ and $j \in [1..n]$, and each partial state $(v' = x, v_{ij} = y)$, where $v'$ is the predecessor of $v_{ij}$ in the causal graph, there is at most one applicable operator for changing the value of $v_{ij}$.*

*Proof.* By inspecting the operators it is easy to see that each pair of operators for $v_{ij}$ have different pre-conditions. The only exception to this rule are operators that do not exist simultaneously due to properties of the CNF formula $F$ (e.g. operators (1) and (2)). ∎

**Lemma 4.5.** *For each partial plan $\Pi$ for $P_{11}(F)$, it holds that*

- $\Pi(v_e) \leq \Pi(v_{kn})$,

- $\Pi(e_1) \leq \Pi(v_e)$, *and*

- $\Pi(e_i) \leq \Pi(e_{i-1})$ *for $i \in [2..2n-1]$.*

*Proof.* Let $v$ be a variable among $v_e, e_1, \ldots, e_{2n-1}$, and let $v'$ be its predecessor in the causal graph. As before, we cannot change the value of $v$ twice without changing the value of $v'$ once in between. If $v \in \{e_1, \ldots, e_{2n-1}\}$, the operator setting $v$ to 1 requires $v' = 1$, and the operator resetting $v$ to 0 requires $v' = 0$. For $v = v_e$, the operator setting $v$ to 1 requires $v'$ to have a value with subscript 0 or 1, while the operator resetting $v$ to 0 requires $v'$ to have a value with subscript $x$. Note that, in either case, we cannot change the value of $v$ in the initial state without first changing the value of $v'$. Thus, $\Pi(v) \leq \Pi(v')$ for each of these variables, as claimed. ∎

We now turn to the problem of finding a plan $\Pi$ that solves $P_{11}(F)$.

**Lemma 4.6.** *Let $\Pi$ be a plan that solves $P_{11}(F)$. Then*

- $\Pi(e_i) \geq 2n - i$ *for $i \in [1..2n-1]$, and*

- $\Pi(v_e) \geq 2n$.

*Proof.* By descending induction on $i$. For $i = 2n - 1$, $\mathsf{goal}(e_{2n-1}) = 1$, so the value of $e_{2n-1}$ has to change at least once from its initial value $\mathsf{init}(e_{2n-1}) = 0$, implying $\Pi(e_{2n-1}) \geq 1 = 2n - (2n - 1)$. For $i \in [1..2n-2]$, assume that $\Pi(e_{i+1}) \geq 2n - (i+1)$ holds by induction. From Lemma 4.5 it follows that $\Pi(e_i) \geq \Pi(e_{i+1}) \geq 2n - (i+1)$. However, since $\mathsf{goal}(e_i) \neq \mathsf{goal}(e_{i+1})$ and since $\Pi$ solves $P_{11}(F)$, it follows that $\Pi(e_i) \neq \Pi(e_{i+1})$. Hence $\Pi(e_i) > \Pi(e_{i+1})$, from which it follows that $\Pi(e_i) \geq 2n - i$, as claimed. The same argument applies to $e_1$ and its predecessor $v_e$, since $\mathsf{goal}(v_e) = 0 \neq 1 = \mathsf{goal}(e_1)$, yielding $\Pi(v_e) \geq 2n$. □

**Definition 4.7.** An admissible plan $\Pi$ for planning problem $P_{11}(F)$ is a partial plan such that $\Pi(s_i) = i$, $\Pi(v_s) = \Pi(v_{11}) = \ldots = \Pi(v_{kn}) = \Pi(v_e) = 2n$, and $\Pi(e_i) = 2n - i$, for each $i \in [1..2n-1]$.

**Lemma 4.8.** *Any plan $\Pi$ that solves $P_{11}(F)$ is admissible.*

*Proof.* By Lemmas 4.3 and 4.5 we have that $\Pi(v_s) \geq \Pi(v_{11}) \geq \cdots \geq \Pi(v_{kn}) \geq \Pi(v_e)$. But, by Lemmas 4.2 and 4.6, all these values are equal to $2n$, since $2n \geq \Pi(v_s)$ and $\Pi(v_e) \geq 2n$. From the proof of Lemma 4.2 we have that $\Pi(s_i) \leq \Pi(s_{i-1}) + 1$, $i \in [2..2n-1]$, and $\Pi(v_s) \leq \Pi(s_{2n-1}) + 1$, which together with Lemma 4.2 and $\Pi(v_s) = 2n$ implies $\Pi(s_i) = i$, $i \in [1..2n-1]$. From the proof of Lemma 4.6 we have that $\Pi(v_e) > \Pi(e_1)$, $\Pi(e_i) > \Pi(e_{i+1})$, $i \in [1..2n-2]$, and $\Pi(e_{2n-1}) \geq 1$, which together with Lemma 4.6 and $\Pi(v_e) = 2n$ implies $\Pi(e_i) = 2n - i$, $i \in [1..2n-1]$. □

Please note that the converse of Lemma 4.8 is not true, that is, not all admissible plans do solve the planning problem $P_{11}(F)$.

As a consequence of Lemma 4.8, to find a plan that solves $P_{11}(F)$ we only need to consider admissible plans. In particular, an admissible plan changes the value of variable $v_s$ exactly $2n$ times, generating a sequence of $2n + 1$ values. Note that the value of $v_s$ always changes from $x$ to either $0$ or $1$, and then back to $x$.

**Definition 4.9.** Let $\Pi$ be an admissible plan, and let $x, m_1, x, \ldots, x, m_n, x$ be the sequence of $2n + 1$ values that variable $v_s$ takes on during the execution of $\Pi$, where $m_j \in \{0, 1\}$ for each $j \in [1..n]$. We use $m_\Pi$ to denote the message $m_1, \ldots, m_n$ induced by $\Pi$, and we use $\sigma_\Pi$ to denote the formula assignment $\sigma_\Pi(x_j) = m_j$ for each $j \in [1..n]$.

As it turns out, the operators that are part of an admissible plan $\Pi$ are completely determined by the message $m_\Pi$ induced by $\Pi$.

**Lemma 4.10.** *Let $\Pi$ be an admissible plan for $P_{11}(F)$ and let $m_\Pi$ be its induced message. The operators in $\Pi$ for changing the value of variable $v_{ij}$, $i \in [1..k]$ and $j \in [1..n]$, as well as the sequence of values that variable $v_{ij}$ takes on during the execution of $\Pi$, are completely determined by $m_\Pi$.*

*Proof.* For each $v \in \{v_{11}, \ldots, v_{kn}\}$, let $v'$ be its causal graph predecessor. From the proof of Lemma 4.3 we know that we cannot change the value of $v$ twice without changing the value of $v'$ in between, and that in the initial state, we have to change the value of $v'$ before we can change the value of $v$. From the definition of admissible we know that $\Pi(v') = \Pi(v) = 2n$. The only way an admissible plan can change the value of $v$ $2n$ times without changing the value of $v'$ more than $2n$ times is to first change the value of $v'$, then $v$, then $v'$, and so on.

Now, from Lemma 4.4 we know that, given a partial state $(v' = x, v = y)$, there is at most one applicable operator for changing the value of $v$. Thus, each time the admissible plan changes the value of $v$ for some value of $v'$, there is at most one operator for doing so. The plan has no choice but to select this operator since it is not allowed to change the value of $v'$ again before changing the value of $v$. Consequently, if the sequence of values taken on by $v'$ is completely determined, the operators for $v$, as well as the sequence of values it takes on, are completely determined also. The proof follows by a double induction on $i$ and $j$, since the sequence of values taken on by $v_s$ (the predecessor of $v_{11}$) is completely determined by the message $m_\Pi$. □

It follows from Lemma 4.10 that the only relevant "degree of freedom" of an admissible plan $\Pi$ is selecting the elements of the message $m_\Pi$, by repeatedly deciding whether to move to $v_s = 0$ or $v_s = 1$ from $v_s = x$. Once $m_\Pi$ has been selected, all other operator choices are forced, else the plan is not admissible. In particular, for each message $m_\Pi$ there is a unique state $s$ such that executing any admissible plan starting from init results in $s$. It remains to determine whether this unique state matches the goal state.

*Remark.* Note that Lemma 4.10 does not mention the operator order of an admissible plan. Indeed, we can change the order of the operators of an admissible plan without making the plan inadmissible. As an example, let $v_1$, $v_2$, and $v_3$ be three consecutive variables in the causal graph, and let $\langle a_1^1, a_2^1, a_3^1, a_1^2, a_2^2, a_3^2 \rangle$ be a subsequence of operators for changing their values, such that $a_i^j$ is the $j$-th operator for changing the value of $v_i$. Then the subsequence $\langle a_1^1, a_2^1, a_1^2, a_3^1, a_2^2, a_3^2 \rangle$ achieves the same result. As long as the partial order $\langle a_i^j, a_{i+1}^j, a_i^{j+1} \rangle$ is respected for each $i$ and $j$, we can change the operator order as we please.

We proceed to determine the sequence of values that variable $v_{ij}$, $i \in [1..k]$ and $j \in [1..n]$, takes on during the execution of an admissible plan $\Pi$ with induced message $m_\Pi$. First, we define the *satisficing index* of clauses, and the *sequence of values* of a plan.

**Definition 4.11.** Let $\Pi$ be an admissible plan with induced message $m_\Pi = m$. For each clause $C_i$, let the *satisficing index* $T_i \in [1..n+1]$ be the smallest number such that $\sigma_\Pi(x_{T_i}) = m_{T_i}$ satisfies $C_i$. If no such number exists, $T_i = n + 1$.

**Definition 4.12.** Let $\Pi$ be an admissible plan. For each clause $C_i$ and each $t \in [1..2n+1]$, let the *sequence of values* $Q_i^t(\Pi)$ be the vector of $n$ values representing, for each variable $v_{ij}$, $j \in [1..n]$, the $t$-th value taken on by $v_{ij}$ during the execution of $\Pi$.

The following lemma is key to understanding the idea behind the reduction for $\mathbb{C}_n^{11}$, since it specifies the sequences of values that an admissible plan induces during its execution.

**Lemma 4.13.** *Let $\sigma$ be an assignment to variables $x_1, \ldots, x_n$ of formula $F$.*

1) **Existence.** *There exists an admissible plan* $\Pi$ *of planning problem* $P_{11}(F)$ *with induced assignment* $\sigma_\Pi = \sigma$.

2) **Claim.** *Let* $Q_i^t$ *be the sequences of values described in Part 3) of this lemma. All admissible plans* $\Pi$ *with* $\sigma_\Pi = \sigma$ *have the same sequences of values* $Q_i^t(\Pi) = Q_i^t$, *for all* $i \in [1..k]$ *and* $t \in [1..2n+1]$.

3) **Sequence of values.** *The sequence of values* $Q_i^t$, *for* $i \in [1..k]$ *and* $t \in [1..2n+1]$, *is as follows.*

    a) *If* $j < T_i$, *then*

$$
\begin{aligned}
Q_i^{2j-1} &= \overbrace{c_x \cdots c_x}^{j-1} \quad a_x \quad \overbrace{a_x \cdots a_x}^{n-j} \\
Q_i^{2j} &= c_{m_j} \cdots c_{m_j} \quad b_{m_j} \quad a_{m_j} \cdots a_{m_j} \\
Q_i^{2j+1} &= c_x \cdots c_x \quad c_x \quad a_x \cdots a_x
\end{aligned}
$$

    b) *If* $j = T_i$, *then*

$$
\begin{aligned}
Q_i^{2j-1} &= \overbrace{c_x \cdots c_x}^{j-1} \quad a_x \quad \overbrace{a_x \cdots a_x}^{n-j} \\
Q_i^{2j} &= c_{m_j} \cdots c_{m_j} \quad g_{m_j} \quad g_{m_j} \cdots g_{m_j} \\
Q_i^{2j+1} &= c_x \cdots c_x \quad g_x \quad g_x \cdots g_x
\end{aligned}
$$

    c) *If* $j > T_i$, *then*

$$
\begin{aligned}
Q_i^{2j-1} &= \overbrace{c_x \cdots c_x}^{T_i-1} \quad \overbrace{g_x \cdots g_x}^{j-T_i} \quad g_x \quad \overbrace{g_x \cdots g_x}^{n-j} \\
Q_i^{2j} &= c_{m_j} \cdots c_{m_j} \quad g_{m_j} \cdots g_{m_j} \quad g_{m_j} \quad g_{m_j} \cdots g_{m_j} \\
Q_i^{2j+1} &= c_x \cdots c_x \quad g_x \cdots g_x \quad g_x \quad g_x \cdots g_x
\end{aligned}
$$

*Proof.* Before proving the lemma, we must check that the definition of $Q_i^t$ given in Part 3 is consistent. This is necessary due to the overlapping of the statements, namely, for every odd $t$ other than 1 and $2n+1$, the sequence $Q_i^t$ is defined twice, once as $Q_i^{2j-1}$ for $j = \lceil \frac{t}{2} \rceil$, and another time as $Q_i^{2j'+1}$ for $j' = \lfloor \frac{t}{2} \rfloor$. However, these sequences of values are well-defined because the definitions of $Q_i^{2j-1}$ and $Q_i^{2j'+1}$ match for any combination of $j$ and $j' = j - 1$, as shown in the following table.

| | $j'$ | $j$ | $Q_i^{2j'+1} = Q_i^{2j-1}$ |
|---|---|---|---|
| $1 < j < T_i$: | Case (a) | Case (a) | $\overbrace{c_x \cdots c_x}^{j'} \overbrace{a_x \cdots a_x}^{n-j'}$ |
| $1 < j = T_i$: | Case (a) | Case (b) | $\overbrace{c_x \cdots c_x}^{j'} \overbrace{a_x \cdots a_x}^{n-j'}$ |
| $j = T_i + 1 \leq n$: | Case (b) | Case (c) | $\overbrace{c_x \cdots c_x}^{T_i-1} \overbrace{g_x \cdots g_x}^{n-T_i+1}$ |
| $T_i + 1 < j \leq n$: | Case (c) | Case (c) | $\overbrace{c_x \cdots c_x}^{T_i-1} \overbrace{g_x \cdots g_x}^{n-T_i+1}$ |

Now, we prove Parts 2 and 3 of the lemma. Assume $\Pi$ is an admissible plan with induced assignment $\sigma_\Pi = \sigma$. The proof proceeds by a double induction on $i$ and $j$. In particular, we prove the validity of the three statements of type $Q_i^{2j-1}, Q_i^{2j}, Q_i^{2j+1}$, assuming that all statements of type $Q_{i'}^t$ (for any $i' < i$ and any $t$) and that all statements of type $Q_i^{2j'-1}, Q_i^{2j'}$ and $Q_i^{2j'+1}$ (for any $j' < j$) already hold. We first prove the validity of $Q_i^{2j-1}$. For $j = 1$, $Q_i^{2j-1} = Q_i^1 = a_x \cdots a_x$ in Cases (a) and (b) corresponds to the initial state of $v_{i1}, \ldots, v_{in}$ (note that Case (c) cannot hold for $j = 1$). When $j > 1$ we know that, since the statements are consistent, $Q_i^{2j-1} = Q_i^{2j'+1}$ for $j' = j - 1$, hence the correctness of $Q_i^{2j-1}$ follows by induction on $j$.

Next, we prove the statements relative to $Q_i^{2j}$ and $Q_i^{2j+1}$. Consider the variable $v'$ that precedes $v_{i1}$ in the causal graph, and values number $2j - 1$, $2j$, and $2j + 1$ it takes on during the execution of $\Pi$. If $i = 1$, then $v' = v_s$ and the values are $x, m_j, x$. If $i > 1$, then $v' = v_{(i-1)n}$ and, by induction on $i$, the values are $a_x, a_{m_j}, a_x$ if $j < T_{i-1}$ and $j < n$; $a_x, b_{m_j}, c_x$ if $n < T_{i-1}$; $a_x, g_{m_j}, g_x$ if $j = T_{i-1}$; or $g_x, g_{m_j}, g_x$ if $j > T_{i-1}$.

The proof is divided into 6 parts, depending on the values of $j$ and $T_i$.

I) $1 = j < T_i$. Consider the following table, where we write $m$ instead of $m_j = m_1$ to simplify the notation.

|          | $v'$                      | $v_{i1}$ | $v_{i2}$ | $\cdots$ | $v_{in}$ |
|----------|---------------------------|----------|----------|----------|----------|
| $2j-1$   | $\{x, a_x, g_x\}$         | $a_x$    | $a_x$    | $\cdots$ | $a_x$    |
| $2j$     | $\{m, a_m, b_m, g_m\}$    | .        | .        | $\cdots$ | .        |
| $2j+1$   | $\{x, a_x, c_x, g_x\}$    | .        | .        | $\cdots$ | .        |

The three rows of the table correspond to values number $2j - 1$, $2j$, and $2j + 1$ of variables $v', v_{i1}, \ldots, v_{in}$. The first column corresponds to the possible values that the predecessor $v'$ of $v_{i1}$ can take on. The first row is given by $Q_i^{2j-1}$, while the second and third rows, to be filled, correspond to $Q_i^{2j}$ and $Q_i^{2j+1}$.

Let $A_{2j}$ be the operator causing the $2j$-th value of $v_{i1}$. According to the previous table, the pre-condition of $A_{2j}$ must be compatible with

$$\langle v' \in \{m_1, a_{m_1}, b_{m_1}, g_{m_1}\}, v_{i1} = a_x \rangle$$

that is, the values of variables $v'$ and $v_{i1}$ when $A_{2j}$ is applied. Since $T_i > 1$, $\sigma_\Pi(x_1) = m_1$ does not satisfy clause $C_i$, so the operator $A_{2j}$ must be one of those labelled (2) and (4) in Table 3. (Only one of these operators is applicable, depending on the value of $m_1$ and whether $v'$ is $v_s$ or $v_{(i-1)n}$.) In either case, the application of $A_{2j}$ causes the value of $v_{i1}$ to become $b_{m_1}$, so we can fill in a blank in the previous table.

|          | $v'$                      | $v_{i1}$      | $v_{i2}$ | $\cdots$ | $v_{in}$ |
|----------|---------------------------|---------------|----------|----------|----------|
| $2j-1$   | $\{x, a_x, g_x\}$         | $a_x$         | $a_x$    | $\cdots$ | $a_x$    |
| $2j$     | $\{m, a_m, b_m, g_m\}$    | $b_m (2,4)$   | .        | $\cdots$ | .        |
| $2j+1$   | $\{x, a_x, c_x, g_x\}$    | .             | .        | $\cdots$ | .        |

In the same way, we can check that $A_{2j+1}$, the operator causing the $(2j + 1)$-th value of $v_{i1}$, must be one of those labelled (7) in Table 3; the new value of $v_{i1}$ is $c_x$. As for the remaining variables, it is easy to check that variables $v_{i2}, \ldots, v_{in}$ become $a_{m_1}$, due to operators of type (14), and then become $a_x$, due to operators of type (18). The table is now complete:

|        | $v'$              | $v_{i1}$    | $v_{i2} \cdots v_{in}$ |
|--------|-------------------|-------------|------------------------|
| $2j-1$ | $\{x, a_x, g_x\}$ | $a_x$       | $a_x \cdots a_x$       |
| $2j$   | $\{m, a_m, b_m, g_m\}$ | $b_m(2,4)$ | $a_m \cdots a_m (14)$ |
| $2j+1$ | $\{x, a_x, c_x, g_x\}$ | $c_x(7)$ | $a_x \cdots a_x (18)$ |

This shows that Case (a) of Lemma 4.13 holds when $j = 1$ and $T_i > 1$.

II) $1 = j = T_i$. The proof is similar to that of Case (I). Since $T_i = 1$, $\sigma_\Pi(x_1) = m_1$ satisfies clause $C_i$. As a result, the admissible operators for causing the $2j$-th value of $v_{i1}$ are now those labelled (1) and (3). In either case, the value of $v_{i1}$ becomes $g_{m_1}$. Consequently, the admissible operators for $v_{i2}, \ldots, v_{in}$ are different from before. This is the resulting table:

|        | $v'$              | $v_{i1}$    | $v_{i2} \cdots v_{in}$ |
|--------|-------------------|-------------|------------------------|
| $2j-1$ | $\{x, a_x, g_x\}$ | $a_x$       | $a_x \cdots a_x$       |
| $2j$   | $\{m, a_m, b_m, g_m\}$ | $g_m(1,3)$ | $g_m \cdots g_m (15)$ |
| $2j+1$ | $\{x, a_x, c_x, g_x\}$ | $g_x(9)$ | $g_x \cdots g_x (21)$ |

III) $1 < j < T_i$. In this case, as in the remaining ones, we just show the resulting table. We always write $m = m_j$. In what follows, we omit the column for $v'$ since its possible values are always the same.

|        | $v_{i1}$    | $v_{i2} \cdots v_{i(j-1)}$ | $v_{ij}$      | $v_{i(j+1)} \cdots v_{in}$ |
|--------|-------------|----------------------------|---------------|-----------------------------|
| $2j-1$ | $c_x$       | $c_x \cdots c_x$           | $a_x$         | $a_x \cdots a_x$            |
| $2j$   | $c_m(5)$    | $c_m \cdots c_m$  (16) | $b_m(11,13)$ | $a_m \cdots a_m$  (14) |
| $2j+1$ | $c_x(8)$    | $c_x \cdots c_x$  (20) | $c_x(19)$    | $a_x \cdots a_x$  (18) |

IV) $1 < j = T_i$.

|        | $v_{i1}$    | $v_{i2} \cdots v_{i(j-1)}$ | $v_{ij}$      | $v_{i(j+1)} \cdots v_{in}$ |
|--------|-------------|----------------------------|---------------|-----------------------------|
| $2j-1$ | $c_x$       | $c_x \cdots c_x$           | $a_x$         | $a_x \cdots a_x$            |
| $2j$   | $c_m(5)$    | $c_m \cdots c_m$  (16) | $g_m(10,12)$ | $g_m \cdots g_m$  (15) |
| $2j+1$ | $c_x(8)$    | $c_x \cdots c_x$  (20) | $g_x(21)$    | $g_x \cdots g_x$  (21) |

V) $1 = T_i < j$.

|        | $v_{i1}$    | $v_{i2} \cdots v_{in}$ |
|--------|-------------|------------------------|
| $2j-1$ | $g_x$       | $g_x \cdots g_x$       |
| $2j$   | $g_m(6)$    | $g_m \cdots g_m(17)$   |
| $2j+1$ | $g_x(9)$    | $g_x \cdots g_x (21)$  |

VI) $1 < T_i < j$.

|        | $v_{i1}$    | $v_{i2} \cdots v_{i(T_i-1)}$ | $v_{iT_i} \cdots v_{in}$ |
|--------|-------------|------------------------------|---------------------------|
| $2j-1$ | $c_x$       | $c_x \cdots c_x$             | $g_x \cdots g_x$          |
| $2j$   | $c_m(5)$    | $c_m \cdots c_m$  (16)  | $g_m \cdots g_m(17)$      |
| $2j+1$ | $c_x(8)$    | $c_x \cdots c_x$  (20)  | $g_x \cdots g_x$  (21) |

It just remains to check that Case (a) of Lemma 4.13 follows from parts (I) and (III), Case (b) from parts (II) and (IV), and Case (c) from parts (V) and (VI). This proves Part 2 and 3 of the lemma.

Finally, note that the existence of an admissible plan $\Pi$ directly follows from the previous discussion, since we have always specified which operators should be used in every situation, and not just assumed their existence. This proves Part 1 of the lemma. □

**Theorem 4.14.** *There exists a plan that solves the planning problem $P_{11}(F)$ if and only if there exists an assignment $\sigma$ that satisfies the CNF formula $F$.*

*Proof.* ⇐: Given an assignment $\sigma$ that satisfies $F$, construct an admissible plan $\Pi$ whose induced formula assignment $\sigma_\Pi$ equals $\sigma$, by choosing the sequence of values of $v_s$ accordingly. It follows that $T_i \leq n$ for each clause $C_i$, since there exists a variable $x_j$ such that $\sigma_\Pi(x_j) = m_j$ satisfies $C_i$. Then, $Q_i^{2n+1}$ has the form indicated in Case (b) or (c) of Lemma 4.13. In either case, the $(2n+1)$-th value of variable $v_{in}$ is $g_x$, as required by the goal state. The plan $\Pi$ thus solves $P_{11}(F)$.

⇒: Let $\Pi$ be a plan that solves the planning problem $P_{11}(F)$. By Lemma 4.8 the plan $\Pi$ is admissible. We show by contradiction that $\sigma = \sigma_\Pi$ satisfies $F$. Assume not. Then there exists a clause $C_i$ not satisfied by $\sigma$, implying $T_i = n + 1$. Since $n < T_i$, the $(2n+1)$-th value of variable $v_{in}$ is $c_x$ according to Case (a) of Lemma 4.13. This contradicts $\Pi$ solving $P_{11}(F)$, since the goal value of $v_{in}$ is not $c_x$ but $g_x$. □

**Proposition 4.15.** *Plan existence for $\mathbb{C}_n^{11}$ is NP-hard.*

*Proof.* The largest variable domains of the planning problem $P_{11}(F)$ are those of variables $v_{11}, \ldots, v_{kn}$, which contain 11 values. The proof follows immediately from the well-known NP-hardness of CNF-SAT, Theorem 4.14, and the fact that we can produce the planning problem $P_{11}(F)$ in polynomial time given the CNF formula $F$. □

### 4.3 Example

We illustrate the reduction using a small example CNF formula $F = (x_1 \lor x_2)$ on one clause and two variables $x_1$ and $x_2$. The variable set of the corresponding planning problem $P_{11}(F)$ is $V = \{s_1, s_2, s_3, v_s, v_{11}, v_{12}, v_e, e_1, e_2, e_3\}$. An admissible plan $\Pi$ can induce any of four different messages $(0,0)$, $(0,1)$, $(1,0)$, and $(1,1)$. Only the message $(0,0)$ corresponds to an assignment that does not satisfy $F$. A plan $\Pi$ that solves $P_{11}(F)$ with induced message $(0,1)$ appears in Table 5. Note that, following execution of the plan, the goal state **goal** $= (v_{12} = g_x, v_e = 0, e_1 = 1, e_2 = 0, e_3 = 1)$ is satisfied as desired; the last value change of each variable appearing in the goal state is marked using boldface.

## 5. $\mathbb{C}_n^5$ Is NP-hard

In this section, we describe a reduction from CNF-SAT to $\mathbb{C}_n^5$. To each CNF formula $F$ we associate a planning problem $P_5(F)$. For each clause $C_i$ and variable $x_j$ of $F$, $P_5(F)$ contains two state variables $v_{ij}^1$, with domain $D(v_{ij}^1) = \{a_x, a_0, a_1, b_x\}$, and $v_{ij}^2$, with domain $D(v_{ij}^2) = \{a_x, a_0, a_1, b_0, b_1\}$. The values $a_0$ and $a_1$ are omitted for $v_{in}^2$, so $D(v_{in}^2) = \{a_x, b_0, b_1\}$. The

$$\langle s_3 = 0, v_s = x; \; v_s = 0 \rangle$$
$$\langle v_s = 0, v_{11} = a_x; \; v_{11} = b_0 \rangle$$
$$\langle v_{11} = b_0, v_{12} = a_x; \; v_{12} = a_0 \rangle$$
$$\langle v_{12} = a_0, v_e = 0; \; v_e = 1 \rangle$$
$$\langle v_e = 1, e_1 = 0; \; e_1 = 1 \rangle$$
$$\langle e_1 = 1, e_2 = 0; \; e_2 = 1 \rangle$$
$$\langle e_2 = 1, e_3 = 0; \; \mathbf{e_3 = 1} \rangle$$
$$\langle s_2 = 0, s_3 = 0; \; s_3 = 1 \rangle$$
$$\langle s_3 = 1, v_s = 0; \; v_s = x \rangle$$
$$\langle v_s = x, v_{11} = b_0; \; v_{11} = c_x \rangle$$
$$\langle v_{11} = c_x, v_{12} = a_0; \; v_{12} = a_x \rangle$$
$$\langle v_{12} = a_x, v_e = 1; \; v_e = 0 \rangle$$
$$\langle v_e = 0, e_1 = 1; \; e_1 = 0 \rangle$$
$$\langle e_1 = 0, e_2 = 1; \; \mathbf{e_2 = 0} \rangle$$
$$\vdots$$

$$\vdots$$
$$\langle s_1 = 0, s_2 = 0; \; s_2 = 1 \rangle$$
$$\langle s_2 = 1, s_3 = 1; \; s_3 = 0 \rangle$$
$$\langle s_3 = 0, v_s = x; \; v_s = 1 \rangle$$
$$\langle v_s = 1, v_{11} = c_x; \; v_{11} = c_1 \rangle$$
$$\langle v_{11} = c_1, v_{12} = a_x; \; v_{12} = g_1 \rangle$$
$$\langle v_{12} = g_1, v_e = 0; \; v_e = 1 \rangle$$
$$\langle v_e = 1, e_1 = 0; \; \mathbf{e_1 = 1} \rangle$$
$$\langle s_1 = 0; \; s_1 = 1 \rangle$$
$$\langle s_1 = 1, s_2 = 1; \; s_2 = 0 \rangle$$
$$\langle s_2 = 0, s_3 = 0; \; s_3 = 1 \rangle$$
$$\langle s_3 = 1, v_s = 1; \; v_s = x \rangle$$
$$\langle v_s = x, v_{11} = c_1; \; v_{11} = c_x \rangle$$
$$\langle v_{11} = c_x, v_{12} = g_1; \; \mathbf{v_{12} = g_x} \rangle$$
$$\langle v_{12} = g_x, v_e = 1; \; \mathbf{v_e = 0} \rangle$$

Table 5: A plan that solves the planning problem $P_{11}(F)$ for the example formula $F$.

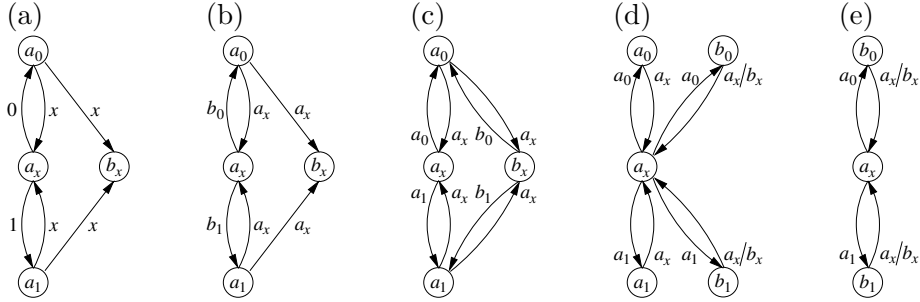

Figure 6: DTGs of (a) $v_{11}^1$, (b) $v_{i1}^1$, $i > 1$, (c) $v_{ij}^1$, $j > 1$, (d) $v_{ij}^2$, $j < n$, (e) $v_{in}^2$.

state variables $s_1, \ldots, s_{2n-1}, v_s, v_e, e_1, \ldots, e_{2n-1}$, as well as their domains and corresponding operators, are the same as before, except the predecessor of $v_e$ is now $v_{kn}^2$.

The initial state on the new state variables is $\mathsf{init}(v_{ij}^1) = \mathsf{init}(v_{ij}^2) = a_x$, $i \in [1..k]$ and $j \in [1..n]$, and the goal state is $\mathsf{goal}(v_{i1}^1) = a_x$, $i \in [1..k]$. Table 6 lists the operators for variables $v_{ij}^1$ and $v_{ij}^2$, $i \in [1..k]$ and $j \in [1..n]$, and Figure 6 shows the corresponding DTGs. Table 6 also lists the new operators for variable $v_e$, which have different pre-conditions now that the predecessor of $v_e$ is $v_{kn}^2$.

## 5.1 Intuition

The reduction for $\mathbb{C}_n^5$ is based on the following idea: instead of using an explicit value to remember that a clause has been satisfied, the goal is to remain in the initial value $a_x$. This way we were able to reduce the size of the variable domains needed for the reduction. Somewhat surprisingly, the new reduction uses fewer total operators than that for $\mathbb{C}_n^{11}$.

| Variable | Ref. | Operator | Qualifier |
|---|---|---|---|
| $v_{11}^1$ | (1) | $\langle v_s = m, v_{11}^1 = a_x; v_{11}^1 = a_m \rangle$ | $m \in \{0,1\}$ |
| | (2) | $\langle v_s = x, v_{11}^1 = a_m; v_{11}^1 = a_x \rangle$ | $m \in \{0,1\}$ |
| | (3) | $\langle v_s = x, v_{11}^1 = a_m; v_{11}^1 = b_x \rangle$ | $m \in \{0,1\}$ |
| $v_{i1}^1$, | (4) | $\langle v_{(i-1)n}^2 = b_m, v_{i1}^1 = a_x; v_{i1}^1 = a_m \rangle$ | $m \in \{0,1\}$ |
| $i \in [2..k]$ | (5) | $\langle v_{(i-1)n}^2 = a_x, v_{i1}^1 = a_m; v_{i1}^1 = a_x \rangle$ | $m \in \{0,1\}$ |
| | (6) | $\langle v_{(i-1)n}^2 = a_x, v_{i1}^1 = a_m; v_{i1}^1 = b_x \rangle$ | $m \in \{0,1\}$ |
| $v_{ij}^1$, | (7) | $\langle v_{i(j-1)}^2 = a_m, v_{ij}^1 = a_x; v_{ij}^1 = a_m \rangle$ | $m \in \{0,1\}$ |
| $i \in [1..k]$, | (8) | $\langle v_{i(j-1)}^2 = a_x, v_{ij}^1 = a_m; v_{ij}^1 = a_x \rangle$ | $m \in \{0,1\}$ |
| $j \in [2..n]$ | (9) | $\langle v_{i(j-1)}^2 = b_m, v_{ij}^1 = b_x; v_{ij}^1 = a_m \rangle$ | $m \in \{0,1\}$ |
| | (10) | $\langle v_{i(j-1)}^2 = a_x, v_{ij}^1 = a_m; v_{ij}^1 = b_x \rangle$ | $m \in \{0,1\}$ |
| $v_{ij}^2$, | (11) | $\langle v_{ij}^1 = a_m, v_{ij}^2 = a_x; v_{ij}^2 = a_m \rangle$ | $m \in \{0,1\}$ |
| $i \in [1..k]$, | (12) | $\langle v_{ij}^1 = a_x, v_{ij}^2 = a_m; v_{ij}^2 = a_x \rangle$ | $m \in \{0,1\}$ |
| $j \in [1..n-1]$ | (13) | $\langle v_{ij}^1 = a_m, v_{ij}^2 = a_x; v_{ij}^2 = b_m \rangle$ | $m \in \{0,1\}$ |
| | (14) | $\langle v_{ij}^1 = a_x, v_{ij}^2 = b_1; v_{ij}^2 = a_x \rangle$ | $x_{n-j+1} \in C_i$ |
| | (15) | $\langle v_{ij}^1 = b_x, v_{ij}^2 = b_1; v_{ij}^2 = a_x \rangle$ | $x_{n-j+1} \notin C_i$ |
| | (16) | $\langle v_{ij}^1 = a_x, v_{ij}^2 = b_0; v_{ij}^2 = a_x \rangle$ | $\overline{x}_{n-j+1} \in C_i$ |
| | (17) | $\langle v_{ij}^1 = b_x, v_{ij}^2 = b_0; v_{ij}^2 = a_x \rangle$ | $\overline{x}_{n-j+1} \notin C_i$ |
| $v_{in}^2$, | (18) | $\langle v_{in}^1 = a_m, v_{in}^2 = a_x; v_{in}^2 = b_m \rangle$ | $m \in \{0,1\}$ |
| $i \in [1..k]$ | (19) | $\langle v_{in}^1 = a_x, v_{in}^2 = b_1; v_{in}^2 = a_x \rangle$ | $x_1 \in C_i$ |
| | (20) | $\langle v_{in}^1 = b_x, v_{in}^2 = b_1; v_{in}^2 = a_x \rangle$ | $x_1 \notin C_i$ |
| | (21) | $\langle v_{in}^1 = a_x, v_{in}^2 = b_0; v_{in}^2 = a_x \rangle$ | $\overline{x}_1 \in C_i$ |
| | (22) | $\langle v_{in}^1 = b_x, v_{in}^2 = b_0; v_{in}^2 = a_x \rangle$ | $\overline{x}_1 \notin C_i$ |
| $v_e$ | | $\langle v_{kn}^2 = b_m, v_e = 0; v_e = 1 \rangle$ | $m \in \{0,1\}$ |
| | | $\langle v_{kn}^2 = a_x, v_e = 1; v_e = 0 \rangle$ | |

Table 6: Operators for variables $v_{ij}^1$, $v_{ij}^2$, and $v_e$, $i \in [1..k]$ and $j \in [1..n]$.

Our reduction for $\mathbb{C}_n^5$ also uses another new idea. In the reduction for $\mathbb{C}_n^{11}$, information was propagated *forward*, i.e., variable $v_{ij}$ changed its value according to the value of its predecessor $v_{i(j-1)}$. The reduction for $\mathbb{C}_n^5$, however, is constructed such that some information is propagated *forward* (in particular, the bits of the message) but other information is propagated *backwards* (the index of the bit we are currently checking). The planning problem is arranged such that a variable $v$ may have several applicable operators, but only one of them satisfies the pre-condition of an applicable action for its successor $v'$. The result is that the value of $v$ at time $t+1$ depends on the value of $v'$ at time $t$.

We explain the planning problem $P_5(F)$ in a bit more detail. Due to the backward propagation mechanism, the bits of the message are checked in reverse order. In other words, $v_{in}$ now checks the first bit, $v_{i(n-1)}$ checks the second bit, and $v_{i1}$ checks the $n$-th bit. The purpose of $v_{ij}^2$ is to check whether the $(n-j+1)$-th bit satisfies the clause $C_i$, whereas the purpose of $v_{ij}^1$ is to inform $v_{i(j-1)}^2$ that the $(n-j+2)$-th bit has arrived. Implicitly, $v_{ij}^1$ also keeps track of whether $C_i$ has been satisfied after the first $(n-j+1)$ bits.

Assume without loss of generality that the message is $0 \cdots 0$. Let us see what happens if the corresponding assignment does not satisfy the clause $C_i$. Upon arrival of the first bit, state variable $v_{in}^2$ has to move to $b_0$. This requires $v_{in}^1 = a_0$ as a pre-condition, which in turn requires state variables $v_{ij}^l$, $j \in [1..n-1]$ and $l \in \{1,2\}$, to be in $a_0$. Next, $v_{in}^2$ has to move back to $a_x$, which requires the pre-condition $v_{in}^1 = b_x$. In turn, this requires state variables $v_{ij}^l$, $j \in [1..n-1]$ and $l \in \{1,2\}$, to be in $a_x$. When $v_{in}^1$ moves again it is from $b_x$ to $a_0$, requiring $v_{i(n-1)}^2 = b_0$ as a pre-condition.

We see that, as long as the clause remains unsatisfied, $v_{ij}^1$ is in $b_x$ following the $(n-j+1)$-th bit. In particular, this means $v_{i1}^1$ is in $b_x$ following the last bit. Assume now that the $(n-j+1)$-th bit satisfies clause $C_i$. When $v_{ij}^2$ moves from $b_0$ to $a_x$, this requires $v_{ij}^1$ to move to $a_x$ instead of $b_x$. From there, there is no way for $v_{i(j-1)}^1$ to be in $b_x$ following the $(n-j+2)$-th bit. In particular, $v_{i1}^1$ will be in $a_x$ following the last bit, satisfying the goal state.

## 5.2 Formal Proof

The proof for $\mathbb{C}_n^5$ is organized much in the same way as that for $\mathbb{C}_n^{11}$. Note that variables $s_1, \ldots, s_{2n-1}, v_s, v_e, e_1, \ldots, e_{2n-1}$ are the same as before, so Lemmas 4.2 and 4.6 still apply to $P_5(F)$. It is easy to check that Lemmas 4.3, 4.5 and 4.8 also hold for $P_5(F)$. However, Lemma 4.4 no longer holds, since several operators share the same preconditions, namely operators (2) and (3), (5) and (6), (8) and (10), and (11) and (13). In spite of this, the operators and sequences of values of an admissible plan $\Pi$ are completely determined by its induced message $m_\Pi$, just as for $P_{11}(F)$ (as shown in Lemma 4.10):

**Lemma 5.1.** *Let $\Pi$ be an admissible plan for $P_5(F)$ and let $m_\Pi$ be its induced message. The operators in $\Pi$ for changing the value of variable $v_{ij}^l$, $i \in [1..k]$, $j \in [1..n]$, and $l \in \{1,2\}$, as well as the sequence of values that variable $v_{ij}^l$ takes on during the execution of $\Pi$, are completely determined by $m_\Pi$.*

*Proof.* First consider variable $v_{11}^1$, and assume without loss of generality that its value is $a_0$. Given $(v_s = x)$, there are two applicable operators for $v_{11}^1$, namely (2), changing its value to $a_x$, and (3), changing its value to $b_x$. At first sight, an admissible plan $\Pi$ can choose either. However, for $\Pi$ to be admissible, it has to change the value of $v_{11}^2$ in between each pair of value changes for $v_{11}^1$. Note that when $v_{11}^1 = a_0$, $v_{11}^2$ can have either of two values, namely $a_0$ or $b_0$. If the value of $v_{11}^2$ is $a_0$, the only admissible operator for $v_{11}^2$ is (12), which has pre-condition $v_{11}^1 = a_x$. Thus, if $\Pi$ changes the value of $v_{11}^1$ to $b_x$ it is no longer admissible, so it has to choose operator (2). If the value of $v_{11}^2$ is $b_0$, the correct choice depends on the CNF formula $F$. If $\overline{x}_n$ satisfies clause $C_1$, the only admissible operator for $v_{11}^2$ is (16) with pre-condition $v_{11}^1 = a_x$, so $\Pi$ should choose operator (2). Otherwise, the only admissible operator for $v_{11}^2$ is (17) with pre-condition is $v_{11}^1 = b_x$, so $\Pi$ should choose operator (3). In either case, the operator choice for $v_{11}^1$ is forced given the value of $v_{11}^2$.

The same reasoning applies to variables $v_{i1}^1$, $i \in [2..k]$, $v_{ij}^1$, $i \in [1..k]$ and $j \in [2..k]$, and $v_{ij}^2$, $i \in [1..k]$ and $j \in [1..n-1]$, and the corresponding operators that share the same pre-conditions. The only degree of freedom of an admissible plan is selecting its induced message $m_\Pi$ by choosing the operators for $v_s$ accordingly. The remaining operator choices and, consequently, sequences of values are completely determined by the induced message $m_\Pi$. □

We now prove a lemma similar to Lemma 4.13, establishing the sequence of values taken on by state variables in $P_5(F)$ during the execution of an admissible plan.

**Definition 5.2.** Let $\Pi$ be an admissible plan for $P_5(F)$. For each clause $C_i$ and each $t \in [1..2n+1]$, let the *sequence of values* $Q_i^t(\Pi)$ be the vector of $2n$ elements representing, for each variable $v_{ij}^l$, $j \in [1..n]$ and $l \in \{1, 2\}$, the $t$-th value taken on by variable $v_{ij}^l$ during the execution of $\Pi$. Let us denote this value by $Q^t(\Pi)[v_{ij}^l]$. We define the *diagonal value* $q_j^i(\Pi)$, for $i \in [1..k]$ and $j \in [1..n]$, as the value $Q^{2j+1}(\Pi)[v_{i(n-j+1)}^1]$.

**Lemma 5.3.** *Let $\sigma$ be an assignment to variables $x_1, \ldots, x_n$ of formula $F$.*

1) **Existence.** *There exists an admissible plan $\Pi$ of planning problem $P_5(F)$ with induced assignment $\sigma_\Pi = \sigma$.*

2) **Claim.** *Let $q_j^i$ be as described in Part 3) of this lemma. All admissible plans $\Pi$ with $\sigma_\Pi = \sigma$ have the same diagonal values $q_j^i(\Pi) = q_j^i$ for each $i \in [1..k]$ and $j \in [1..n]$.*

3) **Diagonal values.** *The diagonal values $q_j^i$, for $i \in [1..k]$ and $j \in [1..n]$, are as follows.*

    *a) If $j < T_i$, then $q_j^i = b_x$.*

    *b) If $j \geq T_i$, then $q_j^i = a_x$.*

*Proof.* Note that, according to Lemma 5.1, not only the diagonal values $q_j^i(\Pi)$, but also the full sequences of values $Q_i^t(\Pi)$, are completely determined for an admissible plan $\Pi$. We have to prove, then, that admissible plans exist for any assignment $\sigma$, as claimed in Part 1, and that the diagonal values match the expression given in Part 3. We prove these two facts by doing a careful, general analysis of the planning problem $P_5(F)$, and then explaining how this analysis implies the lemma. Incidentally, the sequences of values $Q_i^t(\Pi)$ can also be obtained from our analysis; we do not study them because they are not important for our purposes.

Let $\Pi$ be an admissible plan, and let $v = v_{ij}^l$ be some variable of $P_5(F)$. Clearly, the subscript of the $t$-th value $Q^t(\Pi)[v]$ that $v$ takes on depends on the parity of $t$, since all operators affecting $v$ change its subscript from $x$ to $m = \{0,1\}$ and from there back to $x$. Namely, the subscript of $Q^t(\Pi)[v]$ is $x$ if $t = 2p - 1$, and $m$ if $t = 2p$, where $m$ is the $p$-th bit of the message $m_\Pi$.

Now, for some $j \in [2..n-1]$ and $i \in [1..k]$, consider the $t$-th values that variables $v_{ij}^1$, $v_{ij}^2$, $v_{i(j+1)}^1$ take on, for $t = 2p-1, 2p, 2p+1$. The previous observation on the subscripts implies that we (trivially) know something about these values.

|  | $Q^t(\Pi)[v_{ij}^1]$ | $Q^t(\Pi)[v_{ij}^2]$ | $Q^t(\Pi)[v_{i(j+1)}^1]$ |
|---|---|---|---|
| $t = 2p - 1$ | $\{a_x, b_x\}$ | $a_x$ | $\{a_x, b_x\}$ |
| $t = 2p$ | $a_m$ | $\{a_m, b_m\}$ | $a_m$ |
| $t = 2p + 1$ | $\{a_x, b_x\}$ | $a_x$ | $\{a_x, b_x\}$ |

We study how the value $Q^{2p-1}(\Pi)[v_{i(j+1)}^1]$ affects the other values on the diagonal, namely $Q^{2p}(\Pi)[v_{ij}^2]$ and $Q^{2p+1}(\Pi)[v_{ij}^1]$. If $Q^{2p-1}(\Pi)[v_{i(j+1)}^1] = a_x$, then we can check there is only one possible outcome.

| Rule I | $Q^t(\Pi)[v_{ij}^1]$ | | $Q^t(\Pi)[v_{ij}^2]$ | | $Q^t(\Pi)[v_{i(j+1)}^1]$ | |
|---|---|---|---|---|---|---|
| $t = 2p - 1$ | $\{a_x, b_x\}$ | | $a_x$ | | $a_x$ | |
| $t = 2p$ | $a_m$ | | $a_m$ | (11) | $a_m$ | (7) |
| $t = 2p + 1$ | $a_x$ | (8) | $a_x$ | (12) | $\{a_x, b_x\}$ | |

That is, a value of type $a_x$ is propagated along the diagonal to another value $a_x$. We call this Propagation Rule I.

Now we study which are the possible outcomes when $Q^{2p-1}(\Pi)[v_{i(j+1)}^1] = b_x$. In this case, the other values $Q^{2p}(\Pi)[v_{ij}^2]$ and $Q^{2p+1}(\Pi)[v_{ij}^1]$ on the diagonal depend on whether the $p$-th bit $m$ of the message $m_\Pi$ is such that clause $C_i$ is satisfied by $x_{n-j+1} = m$ (c.f. operators (14)–(17) and (18)–(22) in Table 6). If $C_i$ is satisfied by $x_{n-j+1} = m$, it follows that these values must be $b_m$ and $a_x$. This is Propagation Rule II.

| Rule II | $Q^t(\Pi)[v_{ij}^1]$ | | $Q^t(\Pi)[v_{ij}^2]$ | | $Q^t(\Pi)[v_{i(j+1)}^1]$ | |
|---|---|---|---|---|---|---|
| $t = 2p - 1$ | $\{a_x, b_x\}$ | | $a_x$ | | $b_x$ | |
| $t = 2p$ | $a_m$ | | $b_m$ | (13) | $a_m$ | (9) |
| $t = 2p + 1$ | $a_x$ | (8) | $a_x$ | (14,16) | $\{a_x, b_x\}$ | |

On the contrary, if clause $C_i$ is not satisfied, then these values must be $b_m$ and $b_x$. We call this Propagation Rule III.

| Rule III | $Q^t(\Pi)[v_{ij}^1]$ | | $Q^t(\Pi)[v_{ij}^2]$ | | $Q^t(\Pi)[v_{i(j+1)}^1]$ | |
|---|---|---|---|---|---|---|
| $t = 2p - 1$ | $\{a_x, b_x\}$ | | $a_x$ | | $b_x$ | |
| $t = 2p$ | $a_m$ | | $b_m$ | (13) | $a_m$ | (9) |
| $t = 2p + 1$ | $b_x$ | (10) | $a_x$ | (15,17) | $\{a_x, b_x\}$ | |

Finally, let us consider the cases $j = 1$ and $j = n$, which have not been treated in the previous analysis. Note that variables $v_{in}^2$ do not have values of type $a_m$. Also note that variables $v_{i1}^1$ cannot take on value $b_x$ at time $t < 2n + 1$, for then it cannot change further, since the pre-conditions of operators (1)–(3), if $i = 1$, or (4)–(6), if $i \in [2..k]$, are not compatible with $v_{i1}^1 = b_x$. Thus, the only possible outcome for these two variables when $p < n$ is the following.

|  | $Q^t(\Pi)[v_{in}^1]$ | $Q^t(\Pi)[v_{in}^2]$ | | $Q^t(\Pi)[v_{(i+1)1}^1]$ | |
|---|---|---|---|---|---|
| $t = 2p - 1$ | $\{a_x, b_x\}$ | $a_x$ | | $a_x$ | |
| $t = 2p$ | $a_m$ | $b_m$ | (18) | $a_m$ | (4) |
| $t = 2p + 1$ | $\{a_x, b_x\}$ (8; 10) | $a_x$ | (19, 21; 20, 22) | $a_x$ | (5) |

Note that, when $p = n$, the value $Q^{2p+1}(\Pi)[v_{(i+1)1}^1]$ can be either $a_x$ or $b_x$, using operators (5) and (6). The reader can check that a similar analysis applies to variable $v_{11}^1$, where operators (1)–(3) take the role of operators (4)–(6).

Let us summarize the previous analysis in the following table.

| | $v_{i1}^1$ | $v_{i1}^2$ | $v_{i2}^1$ | $\cdots$ | $v_{i(n-1)}^1$ | $v_{i(n-1)}^2$ | $v_{in}^1$ | $v_{in}^2$ |
|---|---|---|---|---|---|---|---|---|
| $t = 1$ | $a_x$ | $a_x$ | $a_x$ | $\cdots$ | $a_x$ | $a_x$ | $a_x$ | $a_x$ |
| $t = 2$ | $a_m$ | | | | | | | $b_m$ |
| $t = 3$ | $a_x$ | | | | | | | $a_x$ |
| $t = 4$ | $a_m$ | | | | | | | $b_m$ |
| $\vdots$ | $\vdots$ | | | | | | | |
| $t = 2n - 2$ | $a_m$ | | | | | | | $b_m$ |
| $t = 2n - 1$ | $a_x$ | | | | | | | $a_x$ |
| $t = 2n$ | $a_m$ | | | | | | | $b_m$ |
| $t = 2n + 1$ | $*$ | | $*$ | $\cdots$ | $*$ | | $*$ | $a_x$ |

The first row in the previous table contains the initial state of the planning problem: all variables are set to $a_x$. The leftmost column and the rightmost column contain the values taken on by variables $v_{i1}^1$ and $v_{in}^2$. Then, the values $b_m$ of the right column are *propagated* along the diagonals using the three propagation rules already discussed: a value of type $a$ yields more values of type $a$ according to Rule I; a value of type $b$ yields a value of type $a$ if the clause is satisfied by Rule II, and of type $b$ if it is not satisfied, by Rule III. The same applies when propagating the values of the first row: since they are all of type $a$, all values of the top-left triangle are of type $a$, according to Rule I. Note also that the longest diagonal coincides with the diagonal values $q_j^i$ of Definition 5.2.

After this discussion we proceed to prove the lemma. Let $\sigma$ be an assignment of formula $F$. The existence of a plan $\Pi$ with $\sigma_\Pi = \sigma$ is implied from the analysis already done on the values $Q^t[v_{ij}^l]$, since we have shown which operators can be used in each case to produce the actual changes of value.

Finally, consider the diagonal values $q_j^i(\Pi)$ for $j = 1, \ldots, n$, that is, the values $Q^3(\Pi)[v_{in}^1]$, $Q^5(\Pi)[v_{i(n-1)}^1]$, $\ldots$, $Q^{2n+1}(\Pi)[v_{i1}^1]$. Let $j < T_i$ as in Case (a), that is, the first $j$ bits of the message $m_\Pi$, when assigned to variables $x_1, \ldots, x_j$, do not satisfy clause $C_i$. Consequently, the diagonal values $q_1^i = Q^3(\Pi)[v_{in}^1]$, $q_2^i = Q^5(\Pi)[v_{i(n-1)}^1]$, $\ldots$, $q_j^i = Q^{2j+1}(\Pi)[v_{i(n+1-j)}^1]$ must all be $b_x$, according to Rule III. On the contrary, if we assume $j \geq T_i$ as in Case (b), then it follows that $q_p^i = b_x$ for $p < T_i$ due to Rule III, that $q_p^i = a_x$ for $p = T_i$ due to Rule II, and that $q_p^i = a_x$ for $j \geq p > T_i$ due to Rule I. □

**Theorem 5.4.** *There exists a valid plan for solving the planning problem $P_5(F)$ if and only if there exists an assignment $\sigma$ that satisfies the CNF formula $F$.*

*Proof.* ⇐: By Lemma 5.3, the existence of an assignment $\sigma$ that satisfies $F$ implies that all admissible plans $\Pi$ with $\sigma_\Pi = \sigma$ satisfy $q_j^i(\Pi) = q_j^i$. Since $T_i \leq n$ for all $i \in [1..k]$, it follows that $q_n^i = a_x$, as required by the goal state of $P_5(F)$. The plan $\Pi$ thus solves $P_5(F)$.

⇒: Let $\Pi$ be a plan solving the planning problem $P_5(F)$. Since Lemma 4.8 holds for $P_5(F)$, the plan $\Pi$ is admissible. We show by contradiction that $\sigma = \sigma_\Pi$ satisfies $F$. Assume not. Then there exists a clause $C_i$ not satisfied by $\sigma$. Thus, Lemma 5.3 implies that $q_j^i(\Pi) = b_x$ for all $j \in [1..n]$. In particular, the value of $v_{i1}^1$ following the execution of $\Pi$ is $b_x$. This contradicts $\Pi$ solving $P_5(F)$, since $b_x$ is different from the goal state $\mathsf{goal}(v_{i1}^1) = a_x$. □

**Proposition 5.5.** *Plan existence for $\mathbb{C}_n^5$ is NP-hard.*

*Proof.* The largest variable domains of the planning problem $P_5(F)$ are those of variables $v_{ij}^2$, $i \in [1..k]$ and $j \in [1..n-1]$, which contain 5 values. The proof follows immediately from the NP-hardness of CNF-SAT, Theorem 5.4, and the fact that we can produce the planning problem $P_5(F)$ in polynomial time given the CNF formula $F$. □

## 6. Discussion

In this paper, we have shown that the problem of determining whether a solution plan exists for planning problems in the class $\mathbb{C}_n^k$ is NP-hard whenever $k \geq 5$. In contrast, Brafman and Domshlak (2003) developed a polynomial-time algorithm for generating plans that solve planning problems in the class $\mathbb{C}_n^2$. What can be said about the intermediate cases, namely $\mathbb{C}_n^k$ for $k \in \{3,4\}$? In what follows, we sketch some arguments for and against tractability of these cases. Although the discussion is mostly based on intuition gained from studying these classes, it might prove helpful for someone trying to determine their complexity.

On one hand, it seems likely to us that plan existence for $\mathbb{C}_n^4$ is also NP-hard. Our reduction for $\mathbb{C}_n^5$ only uses one type of state variable whose domain is larger than 4, namely $v_{ij}^2$. Finding a reduction for $\mathbb{C}_n^4$ seems possible, although it will likely be difficult since the available options become increasingly restricted as the state variable domains get smaller. In particular, we tried but failed to find a reduction for $\mathbb{C}_n^4$.

Domshlak and Dinitz (2001) showed that there exist planning problems in $\mathbb{C}_n^3$ with exponential length minimal solutions. Although this often indicates that a planning class is difficult, it does not imply that plan existence is intractable. This is exemplified by Jonsson and Bäckström (1998) who define a class of planning problems with exponential length minimal solutions but where plan existence could be checked in polynomial time. The present authors (Giménez & Jonsson, 2008a) showed that even plan generation for this particular class could be done in polynomial time, if the resulting plans are given in a compact format such as macros.

A second argument in favor of the hardness of $\mathbb{C}_n^3$ is that there may be multiple ways to transition between two values of a variable. For example, consider a planning problem such that there are two actions for changing the value of a variable $v$ from 0 to 1, namely $a = \langle v' = 0, v = 0; v = 1 \rangle$ and $a' = \langle v' = 1, v = 0; v = 1 \rangle$. Since variables can have 3 values, it is possible that neither $v' = 0$ nor $v' = 1$ hold in the current state. A planner would thus have to *choose* whether to satisfy $v' = 0$ or $v' = 1$. In contrast, for $\mathbb{C}_n^2$ the same two actions could be replaced by a single action $\langle v = 0; v = 1 \rangle$ since one of $a$ and $a'$ is always applicable. As a consequence, even if the minimal plan length is bounded for a planning problem in $\mathbb{C}_n^3$, there may be exponentially many plans of that length (in fact, this is the main idea behind our reductions).

Another observation regards the number of possible domain transition graphs for each state variable. For each $k \geq 2$, it is possible to show that a state variable in $\mathbb{C}_n^k$ may have $2^{k^2(k-1)}$ distinct domain transition graphs. In other words, the number of graphs grows exponentially in $k$. In particular, while state variables in $\mathbb{C}_n^2$ can only have $2^4 = 16$ distinct graphs, the same number for $\mathbb{C}_n^3$ is $2^{18}$. Although a large number of possibilities does not guarantee hardness, it is clear that the expressive power of $\mathbb{C}_n^3$ is much higher than that of $\mathbb{C}_n^2$.

The evidence provided above suggests that $\mathbb{C}_n^3$ is significantly harder than $\mathbb{C}_n^2$. However, we are not sure that $\mathbb{C}_n^3$ is hard enough to be intractable. State variables with just three values do not lend themselves well to the type of reduction we have presented, since just propagating the message requires three values. If there is such a reduction for $\mathbb{C}_n^3$, the idea underlying it may not be the message-passing mechanism we have exploited. On the other hand, maybe there is some way to determine plan existence of $\mathbb{C}_n^3$ in polynomial time. Such an algorithm would take into consideration the multiple (but finite) combinations of domain transition graphs of three values, as well as any inherent structure of the graphs. We know that the expressive power of domain transition graphs of 5 values is just too large to handle in polynomial time; maybe this is not the case when using just 3 values.

## Acknowledgments

This work was partially funded by APIDIS and MEC grant TIN2006-15387-C03-03.

## Appendix A. $\mathbb{C}_n^7$ Is NP-hard

In this appendix, we describe how to modify the reduction for $\mathbb{C}_n^{11}$ so that the resulting planning problem, which we call $P_7(F)$, only needs variable domains of size 7. This reduction previously appeared in a conference paper (Giménez & Jonsson, 2008b), but without proof. The main idea of the reduction is the same, but the construction used to check if the assignment $\sigma_\Pi$ satisfies a clause $C_i$ is more involved. Previously, we used $n$ variables $\{v_{ij}\}_{j \in [1..n]}$ whose role was, essentially, to check whether the $j$-th bit $\sigma_\Pi(x_j)$ of the propagated message satisfies $C_i$. In the modified reduction, each variable $v_{ij}$ is replaced by three variables $v_{ij}^1$, $v_{ij}^2$, and $v_{ij}^3$, that collectively play the same role. The variables $s_1, \ldots, s_{2n-1}, v_s, v_e, e_1, \ldots, e_{2n-1}$, as well as their domains and corresponding operators, are the same as before, except the predecessor of $v_e$ is now $v_{kn}^3$.

The domains of these new variables are $D(v_{ij}^1) = D(v_{ij}^3) = \{a_x, a_0, a_1, b_x, b_0, b_1, g_x\}$ and $D(v_{ij}^2) = \{g_x, g_0, g_1, a_x, a_0, a_1, b_x\}$ for each $i \in [1..k]$, $j \in [1..n]$. The initial state on these variables is $\mathsf{init}(v_{ij}^1) = \mathsf{init}(v_{ij}^2) = \mathsf{init}(v_{ij}^3) = a_x$, $i \in [1..k]$ and $j \in [1..n]$, and the goal state is $\mathsf{goal}(v_{in}^2) = g_x$, $i \in [1..k]$. Table 7 shows the operators for variables $v_{ij}^1$, $i \in [1..k]$ and $j \in [1..n]$, and Table 8 shows the operators for variables $v_{ij}^2$ and $v_{ij}^3$, $i \in [1..k]$ and $j \in [1..n]$. Figures 7 and 8 shows the corresponding domain transition graphs. Table 8 also shows the new operators for variable $v_e$, which have different pre-conditions now that the predecessor of $v_e$ is $v_{kn}^3$.

| Variable | Ref. | Operator | Qualifier |
|---|---|---|---|
| $v_{11}^1$ | (1) | $\langle v_s = 1, v_{11}^1 = a_x;\, v_{11}^1 = g_x \rangle$ | $x_1 \in C_1$ |
| | (2) | $\langle v_s = 1, v_{11}^1 = a_x;\, v_{11}^1 = b_1 \rangle$ | $x_1 \notin C_1$ |
| | (3) | $\langle v_s = 0, v_{11}^1 = a_x;\, v_{11}^1 = g_x \rangle$ | $\overline{x}_1 \in C_1$ |
| | (4) | $\langle v_s = 0, v_{11}^1 = a_x;\, v_{11}^1 = b_0 \rangle$ | $\overline{x}_1 \notin C_1$ |
| | (5) | $\langle v_s = m, v_{11}^1 = g_x;\, v_{11}^1 = b_m \rangle$ | $m \in \{0,1\}$ |
| | (6) | $\langle v_s = m, v_{11}^1 = b_x;\, v_{11}^1 = b_m \rangle$ | $m \in \{0,1\}$ |
| | (7) | $\langle v_s = x, v_{11}^1 = b_m;\, v_{11}^1 = b_x \rangle$ | $m \in \{0,1\}$ |
| $v_{i1}^1$, | (1) | $\langle v_{(i-1)n}^3 \in \{a_1, b_1\}, v_{i1}^1 = a_x;\, v_{i1}^1 = g_x \rangle$ | $x_1 \in C_i$ |
| $i \in [2..k]$ | (2) | $\langle v_{(i-1)n}^3 \in \{a_1, b_1\}, v_{i1}^1 = a_x;\, v_{i1}^1 = b_1 \rangle$ | $x_1 \notin C_i$ |
| | (3) | $\langle v_{(i-1)n}^3 \in \{a_0, b_0\}, v_{i1}^1 = a_x;\, v_{i1}^1 = g_x \rangle$ | $\overline{x}_1 \in C_i$ |
| | (4) | $\langle v_{(i-1)n}^3 \in \{a_0, b_0\}, v_{i1}^1 = a_x;\, v_{i1}^1 = b_0 \rangle$ | $\overline{x}_1 \notin C_i$ |
| | (5) | $\langle v_{(i-1)n}^3 \in \{a_m, b_m\}, v_{i1}^1 = g_x;\, v_{i1}^1 = b_m \rangle$ | $m \in \{0,1\}$ |
| | (6) | $\langle v_{(i-1)n}^3 \in \{a_m, b_m\}, v_{i1}^1 = b_x;\, v_{i1}^1 = b_m \rangle$ | $m \in \{0,1\}$ |
| | (7) | $\langle v_{(i-1)n}^3 \in \{a_x, b_x\}, v_{i1}^1 = b_m;\, v_{i1}^1 = b_x \rangle$ | $m \in \{0,1\}$ |
| $v_{ij}^1$, | (8) | $\langle v_{i(j-1)}^3 = b_1, v_{ij}^1 = a_x;\, v_{ij}^1 = g_x \rangle$ | $x_j \in C_i$ |
| $i \in [1..k]$, | (9) | $\langle v_{i(j-1)}^3 = b_1, v_{ij}^1 = a_x;\, v_{ij}^1 = b_1 \rangle$ | $x_j \notin C_i$ |
| $j \in [2..n]$ | (10) | $\langle v_{i(j-1)}^3 = b_0, v_{ij}^1 = a_x;\, v_{ij}^1 = g_x \rangle$ | $\overline{x}_j \in C_i$ |
| | (11) | $\langle v_{i(j-1)}^3 = b_0, v_{ij}^1 = a_x;\, v_{ij}^1 = b_0 \rangle$ | $\overline{x}_j \notin C_i$ |
| | (12) | $\langle v_{i(j-1)}^3 = g_x, v_{ij}^1 = a_x;\, v_{ij}^1 = g_x \rangle$ | |
| | (13) | $\langle v_{i(j-1)}^3 = a_m, v_{ij}^1 = a_x;\, v_{ij}^1 = a_m \rangle$ | $m \in \{0,1\}$ |
| | (14) | $\langle v_{i(j-1)}^3 = b_m, v_{ij}^1 = g_x;\, v_{ij}^1 = b_m \rangle$ | $m \in \{0,1\}$ |
| | (15) | $\langle v_{i(j-1)}^3 = b_m, v_{ij}^1 = b_x;\, v_{ij}^1 = b_m \rangle$ | $m \in \{0,1\}$ |
| | (16) | $\langle v_{i(j-1)}^3 \in \{a_x, b_x\}, v_{ij}^1 = a_m;\, v_{ij}^1 = a_x \rangle$ | $m \in \{0,1\}$ |
| | (17) | $\langle v_{i(j-1)}^3 = b_x, v_{ij}^1 = b_m;\, v_{ij}^1 = b_x \rangle$ | $m \in \{0,1\}$ |

Table 7: Operators for variables $v_{ij}^1$, $i \in [1..k]$ and $j \in [1..n]$.

## A.1 Intuition

The intuition behind the reduction for $\mathbb{C}_n^7$ is largely the same as that for $\mathbb{C}_n^{11}$. The planning problem $P_7(F)$ corresponding to a CNF formula $F$ consists of three parts, the first and third being identical to those of $P_{11}(F)$. Thus, the difference lies in the second part. Recall that in the reduction for $\mathbb{C}_n^{11}$, for each clause $C_i$ and each variable $x_j$ of $F$, the planning problem $P_{11}(F)$ contains a state variable $v_{ij}$ that performs the following functions:

1. Propagate the message $m$ generated by $v_s$.

| Variable | Ref. | Operator | Qualifier |
|---|---|---|---|
| $v_{ij}^2$, | (18) | $\langle v_{ij}^1 \in \{a_m, b_m\}, v_{ij}^2 = a_x; v_{ij}^2 = a_m \rangle$ | $m \in \{0,1\}$ |
| $i \in [1..k]$, | (19) | $\langle v_{ij}^1 = g_x, v_{ij}^2 = a_x; v_{ij}^2 = g_x \rangle$ | |
| $j \in [1..n]$ | (20) | $\langle v_{ij}^1 = b_m, v_{ij}^2 = g_x; v_{ij}^2 = g_m \rangle$ | $m \in \{0,1\}$ |
| | (21) | $\langle v_{ij}^1 = b_m, v_{ij}^2 = b_x; v_{ij}^2 = a_m \rangle$ | $m \in \{0,1\}$ |
| | (22) | $\langle v_{ij}^1 = a_x, v_{ij}^2 = a_m; v_{ij}^2 = a_x \rangle$ | $m \in \{0,1\}$ |
| | (23) | $\langle v_{ij}^1 = b_x, v_{ij}^2 = a_m; v_{ij}^2 = b_x \rangle$ | $m \in \{0,1\}$ |
| | (24) | $\langle v_{ij}^1 = b_x, v_{ij}^2 = g_m; v_{ij}^2 = g_x \rangle$ | $m \in \{0,1\}$ |
| $v_{ij}^3$, | (25) | $\langle v_{ij}^2 = a_m, v_{ij}^3 = a_x; v_{ij}^3 = a_m \rangle$ | $m \in \{0,1\}$ |
| $i \in [1..k]$, | (26) | $\langle v_{ij}^2 = g_x, v_{ij}^3 = a_x; v_{ij}^3 = g_x \rangle$ | |
| $j \in [1..n]$ | (27) | $\langle v_{ij}^2 = g_m, v_{ij}^3 = g_x; v_{ij}^3 = b_m \rangle$ | $m \in \{0,1\}$ |
| | (28) | $\langle v_{ij}^2 \in \{a_m, g_m\}, v_{ij}^3 = b_x; v_{ij}^3 = b_m \rangle$ | $m \in \{0,1\}$ |
| | (29) | $\langle v_{ij}^2 = a_x, v_{ij}^3 = a_m; v_{ij}^3 = a_x \rangle$ | $m \in \{0,1\}$ |
| | (30) | $\langle v_{ij}^2 = b_x, v_{ij}^3 = a_m; v_{ij}^3 = b_x \rangle$ | $m \in \{0,1\}$ |
| | (31) | $\langle v_{ij}^2 \in \{b_x, g_x\}, v_{ij}^3 = b_m; v_{ij}^3 = b_x \rangle$ | $m \in \{0,1\}$ |
| $v_e$ | | $\langle v_{kn}^3 \in \{a_0, a_1, b_0, b_1\}, v_e = 0; v_e = 1 \rangle$ | |
| | | $\langle v_{kn}^3 \in \{a_x, b_x\}, v_e = 1; v_e = 0 \rangle$ | |

Table 8: Operators for variables $v_{ij}^2$, $v_{ij}^3$, and $v_e$, $i \in [1..k]$ and $j \in [1..n]$.

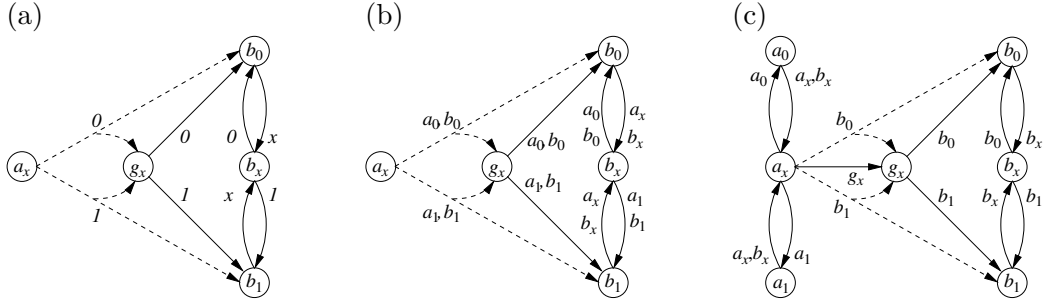

Figure 7: DTGs of (a) $v_{11}^1$, (b) $v_{i1}^1$ for $i \in [2..k]$, (c) $v_{ij}^1$ for $i \in [1..k]$, $j \in [2..n]$.

2. Check whether the assignment to $x_j$ (the $j$-th bit of $m$) satisfies the clause $C_i$.

3. Remember whether $C_i$ was satisfied by the assignment to some $x_l$, $l \leq j$.

4. If $j < n$ and $C_i$ has been satisfied, propagate this fact.

5. If $j < n$, let $v_{i(j+1)}$ know when the $(j + 1)$-th bit of the message has arrived.

The first and fourth function is to propagate information and thus has to be performed by all state variables if information is not to be lost. However, the other functions can be performed by different state variables. The idea behind the reduction for $\mathbb{C}_n^7$ is to split $v_{ij}$ into three variables: $v_{ij}^1$, that performs the second function, $v_{ij}^2$, that performs the third, and $v_{ij}^3$, that performs the fifth.

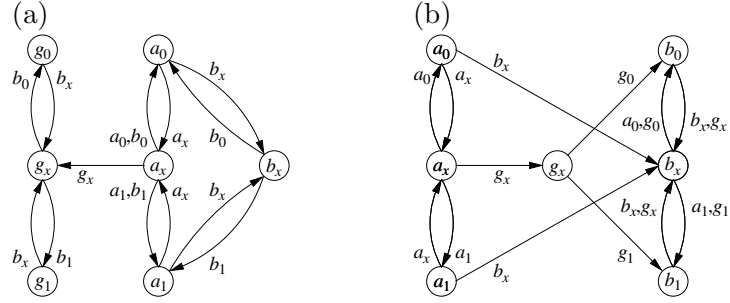

Figure 8: DTGs of (a) $v_{ij}^2$ and (b) $v_{ij}^3$ for $i \in [1..k], j \in [1..n]$.

Just as before, the message $m$ is propagated using the subscripts of values in the domains of state variables. When the $j$-th bit $m_j$ of the message arrives, state variable $v_{ij}^1$ moves from $a_x$ to $g_x$ if the assignment $\sigma(x_j) = m_j$ satisfies $C_i$, and to $b_{m_j}$ otherwise. If $v_{ij}^1$ moves to $g_x$, it is forced to move to $b_{m_j}$ next, forgetting that $C_i$ was satisfied. However, while the value of $v_{ij}^1$ is $g_x$, all subsequent state variables for $C_i$ can also move to $g_x$, propagating the fact that $C_i$ has been satisfied. Consequently, state variable $v_{in}^2$ is able to remember that $C_i$ has been satisfied by remaining within the subdomain $\{g_0, g_1, g_x\}$.

If $\sigma(x_j) = m_j$ does not satisfy $C_i$, $v_{ij}^1$ moves to $b_{m_j}$, causing $v_{ij}^2$ and $v_{ij}^3$ to move to $a_{m_j}$. From there, $v_{ij}^1$, $v_{ij}^2$, and $v_{ij}^3$ all move to $b_x$. When the next bit arrives, $v_{ij}^1$ moves to $b_0$ ($b_1$), causing $v_{ij}^2$ to move to $a_0$ ($a_1$) and $v_{ij}^3$ to move to $b_0$ ($b_1$). This indicates to $v_{i(j+1)}^1$ that the $(j+1)$-th bit has arrived, causing it to act accordingly. Just as before, the operators are defined such that $v_{i1}^1$ always reacts to the first bit for each clause $C_i$.

## A.2 Formal Proof

Since variables $s_1, \ldots, s_{2n-1}, v_s, v_e, e_1, \ldots, e_{2n-1}$ are the same as before, Lemmas 4.2 and 4.6 both apply to $P_7(F)$. However, Lemma 4.3 is violated since it is sometimes possible to change the value of a variable twice without changing the value of its predecessor (e.g. using operators (1) and (5)). Consequently, Lemma 4.8, which states that all plans that solve $P_{11}(F)$ are admissible, no longer holds for $P_7(F)$.

To prove equivalent lemmas for $P_7(F)$, we redefine $\Pi(v_{ij}^l)$ for variables in the middle of the causal graph:

**Definition A.1.** Given a partial plan $\Pi$ and variable $v_{ij}^l$, $i \in [1..k]$, $j \in [1..n]$, and $l \in \{1, 2, 3\}$, let $\Pi(v_{ij}^l)$ be the number of subscript changes of $v_{ij}^l$ during the execution of $\Pi$.

**Lemma A.2.** For each partial plan $\Pi$ for $P_7(F)$ and each $v_{ij}^l$, $i \in [1..k]$, $j \in [1..n]$, and $l \in \{1, 2, 3\}$, it holds that $\Pi(v_{ij}^l) \leq \Pi(v')$, where $v'$ is the predecessor of $v_{ij}^l$ in the causal graph.

*Proof.* Follows immediately from inspection of the operators for $v_{ij}^l$. Each operator that changes the subscript of $v_{ij}^l$ to $z \in \{0, 1, x\}$ has a pre-condition on $v'$ with subscript $z$ (or value $z$ in the case of $v_{11}^1$ and its predecessor $v_s$). There are operators for changing the value of $v_{ij}^1$ to $g_x$ that have a pre-condition on $v'$ with a subscript (or value) different from $x$, but these operators do not change the subscript of $v_{ij}^1$ since their pre-condition on $v_{ij}^1$ is $a_x$. □

**Lemma A.3.** *For each partial plan $\Pi$ for $P_7(F)$, $\Pi(v_e) \leq \Pi(v_{kn}^3)$.*

*Proof.* Note that $\Pi(v_e)$ still denotes the number of value changes of $v_e$, while $\Pi(v_{kn}^3)$ denotes the number of subscript changes of $v_{kn}^3$. Each time we change the value of $v_e$ we need to change the subscript of $v_{kn}^3$ in between. In addition, the first value change of $v_e$ requires a subscript for $v_{kn}^3$ different from that in the initial state. Thus, $\Pi(v_e) \leq \Pi(v_{kn}^3)$. $\square$

**Definition A.4.** An admissible plan $\Pi$ for planning problem $P_7(F)$ is a partial plan such that $\Pi(s_i) = i$, $\Pi(v_s) = \Pi(v_{11}^1) = \ldots = \Pi(v_{kn}^3) = \Pi(v_e) = 2n$, and $\Pi(e_i) = 2n - i$, for each $i \in [1 .. 2n-1]$.

**Lemma A.5.** *Any plan $\Pi$ that solves the planning problem $P_7(F)$ is admissible.*

*Proof.* By Lemmas A.2 and A.3 we have that $\Pi(v_s) \geq \Pi(v_{11}^1) \geq \cdots \geq \Pi(v_{kn}^3) \geq \Pi(v_e)$. We can now use Lemmas 4.2 and 4.6 and apply the same reasoning as in the proof of Lemma 4.8. $\square$

In other words, an admissible plan has to change the subscript of each $v_{ij}^l$ exactly $2n$ times, although it can change the value of $v_{ij}^l$ an extra time by moving through $g_x$. However, even with the new definition of $\Pi(v_{ij}^l)$, we cannot prove an equivalent of Lemma 4.10 for $P_7(F)$, since a variable $v_{ij}^l$, $l \in \{1, 2\}$, can choose not to follow its predecessor to $g_x$ without making the plan inadmissible. Consequently, the sequences of values $Q_i^t(\Pi)$ of an admissible plan $\Pi$ are no longer completely determined by the induced message $m_\Pi$. Nevertheless, we can still prove a lemma similar to Lemma 4.13.

**Definition A.6.** Let $\Pi$ be an admissible plan. For each clause $C_i$ and each $t \in [1 .. 2n+1]$, let the *sequence of values* $Q_i^t(\Pi)$ be the vector of $3n$ elements representing, for each variable $v_{ij}^l$, $j \in [1 .. n]$ and $l \in \{1, 2, 3\}$, the first value following the $(t-1)$-th subscript change of $v_{ij}^l$ during the execution of $\Pi$.

**Lemma A.7.** *Let $\sigma$ be an assignment of variables $x_1, \ldots, x_n$ of formula $F$.*

1) *Existence. There exists an admissible plan $\Pi$ of planning problem $P_7(F)$ with induced assignment $\sigma_\Pi = \sigma$.*

2) *Claim. Let $Q_i^t$ be the sequences of values described in Part 3) of this lemma. If $\sigma$ satisfies $F$, then there exists an admissible plan $\Pi$ with $\sigma_\Pi = \sigma$ such that $Q_i^t(\Pi) = Q_i^t$, for all $t \in [1 .. 2n+1]$ and $i \in [1 .. k]$. If $\sigma$ does not satisfy clause $C_i$, then all admissible plans $\Pi$ with $\sigma_\Pi = \sigma$ have $Q_i^t(\Pi) = Q_i^t$, for all $t \in [1 .. 2k + 1]$.*

3) *Sequence of values. The sequence of values $Q_i^t$, for $i \in [1 .. k]$ and $t \in [1 .. 2n+1]$, is as follows.*

   a) *If $j < T_i$, then*

$$
\begin{array}{rccc}
Q_i^{2j-1} = & \overbrace{b_x b_x b_x \cdots b_x b_x b_x}^{j-1} & a_x a_x a_x & \overbrace{a_x a_x a_x \cdots a_x a_x a_x}^{n-j} \\
Q_i^{2j} = & b_m a_m b_m \cdots b_m a_m b_m & b_m a_m a_m & a_m a_m a_m \cdots a_m a_m a_m \\
Q_i^{2j+1} = & b_x b_x b_x \cdots b_x b_x b_x & b_x b_x b_x & a_x a_x a_x \cdots a_x a_x a_x
\end{array}
$$

*b) If $j = T_i$, then*

$$Q_i^{2j-1} = \overbrace{b_x b_x b_x \cdots b_x b_x b_x}^{j-1} \quad a_x a_x a_x \quad \overbrace{a_x a_x a_x \cdots a_x a_x a_x}^{n-j}$$
$$Q_i^{2j} = b_m a_m b_m \cdots b_m a_m b_m \quad b_m g_m b_m \quad b_m g_m b_m \cdots b_m g_m b_m$$
$$Q_i^{2j+1} = b_x b_x b_x \cdots b_x b_x b_x \quad b_x g_x b_x \quad b_x g_x b_x \cdots b_x g_x b_x$$

*c) If $j > T_i$, then*

$$Q_i^{2j-1} = \overbrace{b_x b_x b_x \cdots b_x b_x b_x}^{T_i-1} \quad \overbrace{b_x g_x b_x \cdots b_x g_x b_x}^{j-T_i} \quad b_x g_x b_x \quad \overbrace{b_x g_x b_x \cdots b_x g_x b_x}^{n-j}$$
$$Q_i^{2j} = b_m a_m b_m \cdots b_m a_m b_m \; b_m g_m b_m \cdots b_m g_m b_m \; b_m g_m b_m \; b_m g_m b_m \cdots b_m g_m b_m$$
$$Q_i^{2j+1} = b_x b_x b_x \cdots b_x b_x b_x \quad b_x g_x b_x \cdots b_x g_x b_x \quad b_x g_x b_x \quad b_x g_x b_x \cdots b_x g_x b_x$$

*Proof.* Note the similarity of this lemma with Lemma 4.13. As before, we must show that there are operators, this time in Tables 7 and 8, whose post-conditions equal the values given by $Q_i^{2j-1}$, $Q_i^{2j}$ and $Q_i^{2j+1}$. Again, we must check for consistency in the statements of $Q_i^{2j-1}$ and $Q_i^{2j'+1}$ with $j' = j - 1$. This implies, as in Lemma 4.13, that the statements for $Q_i^{2j-1}$ are valid, due to the initial state being $a_x \cdots a_x$ and by induction on $j$. It just remains to show that the statements for $Q_i^{2j}$ and $Q_i^{2j+1}$ are also valid.

The proof is divided into the same six parts as that of Lemma 4.13. Note that, in contrast to that lemma, here we aim to show that, when $\sigma$ satisfies $F$, there exists an admissible plan with given $Q_i^t$, not that all admissible plans have this form. This is because sometimes during the execution of the plan more than one operator could be chosen, and the resulting plan would still be admissible. In the tables that follow, which are alike to those in the proof of Lemma 4.13, we only indicate the operator choice that leads to the desired $Q_i^t$, and we use boldface to remark that these operators are not "forced". We add an extra row to the tables to indicate that sometimes we need to apply two operators for each variable before changing its subscript. These disparities with respect to Lemma 4.13 only occur in parts II and IV of the proof, which require $T_i \leq n$, that is, $\sigma$ satisfying clause $C_i$, for some fixed $i$. Thus, when $\sigma$ does not satisfy clause $C_i$, all admissible plans $\Pi$ have the same sequences of values $Q_i^t$ for each $t \in [1..2n+1]$.

I) $1 = j < T_i$.

| | $v_{i1}^1 v_{i1}^2 v_{i1}^3$ | $v_{ik}^1 v_{ik}^2 v_{ik}^3 \; \vert k \in [2..n]$ |
|---|---|---|
| $2j-1$ | $a_x a_x a_x$ | $a_x a_x a_x$ |
| $2j$ | $b_m a_m a_m (2, 4; 18; 25)$ | $a_m a_m a_m \; (13; 18; 25)$ |
| $2j+1$ | $b_x b_x b_x \; (7; 23; 30)$ | $a_x a_x a_x \; (16; 22; 29)$ |

II) $1 = j = T_i$.

| | $v_{i1}^1 v_{i1}^2 v_{i1}^3$ | $v_{ik}^1 v_{ik}^2 v_{ik}^3 \; \vert k \in [2..n]$ |
|---|---|---|
| $2j-1$ | $a_x a_x a_x$ | $a_x a_x a_x$ |
| | $g_x g_x g_x (1, 3; \mathbf{19}; 26)$ | $g_x g_x g_x \; (\mathbf{12}; \mathbf{19}; 26)$ |
| $2j$ | $b_m g_m b_m (5; 20; 27)$ | $b_m g_m b_m \; (14; 20; 27)$ |
| $2j+1$ | $b_x g_x b_x (7; 24; 31)$ | $b_x g_x b_x \; (17; 24; 31)$ |

III) $1 < j < T_i$.

| | $v_{ik}^1 v_{ik}^2 v_{ik}^3 \ \vert k \in [1..j-1]$ | $v_{ij}^1 v_{ij}^2 v_{ij}^3$ | $v_{ik}^1 v_{ik}^2 v_{ik}^3 \ \vert k \in [j+1..n]$ |
|---|---|---|---|
| $2j-1$ | $b_x b_x b_x$ | $a_x a_x a_x$ | $a_x a_x a_x$ |
| $2j$ | $b_m a_m b_m$ $(6, 15; 21; 28)$ | $b_m a_m a_m (9, 11; 18; 25)$ | $a_m a_m a_m$ $(13; 18; 25)$ |
| $2j+1$ | $b_x b_x b_x$ $(7, 17; 23; 31)$ | $b_x b_x b_x$ $(17; 23; 30)$ | $a_x a_x a_x$ $(16; 22; 29)$ |

IV) $1 < j = T_i$.

| | $v_{ik}^1 v_{ik}^2 v_{ik}^3 \ \vert k \in [1..j-1]$ | $v_{ij}^1 v_{ij}^2 v_{ij}^3$ | $v_{ik}^1 v_{ik}^2 v_{ik}^3 \ \vert k \in [j+1..n]$ |
|---|---|---|---|
| $2j-1$ | $b_x b_x b_x$ | $a_x a_x a_x$ | $a_x a_x a_x$ |
| $2j$ | $b_m a_m b_m$ $(6, 15; 21; 28)$ | $g_x g_x g_x$ $(8, 10; \mathbf{19}; 26)$ | $g_x g_x g_x$ $(\mathbf{12}; \mathbf{19}; 26)$ |
| | $b_m a_m b_m$ | $b_m g_m b_m (14; 20; 27)$ | $b_m g_m b_m$ $(14; 20; 27)$ |
| $2j+1$ | $b_x b_x b_x$ $(7, 17; 23; 31)$ | $b_x g_x b_x$ $(17; 24; 31)$ | $b_x g_x b_x$ $(17; 24; 31)$ |

V) $1 = T_i < j$.

| | $v_{i1}^1 v_{i1}^2 v_{i1}^3$ | $v_{ik}^1 v_{ik}^2 v_{ik}^3 \ \vert k \in [2..n]$ |
|---|---|---|
| $2j-1$ | $b_x g_x b_x$ | $b_x g_x b_x$ |
| $2j$ | $b_m g_m b_m (6; 20; 28)$ | $b_m g_m b_m$ $(15; 20; 28)$ |
| $2j+1$ | $b_x g_x b_x (7; 24; 31)$ | $b_x g_x b_x$ $(17; 24; 31)$ |

VI) $1 < T_i < j$.

| | $v_{ik}^1 v_{ik}^2 v_{ik}^3 \ \vert k \in [1..T_i-1]$ | $v_{ik}^1 v_{ik}^2 v_{ik}^3 \ \vert k \in [T_i..n]$ |
|---|---|---|
| $2j-1$ | $b_x b_x b_x$ | $b_x g_x b_x$ |
| $2j$ | $b_m a_m b_m$ $(6, 15; 21; 28)$ | $b_m g_m b_m$ $(15; 20; 28)$ |
| $2j+1$ | $b_x b_x b_x$ $(7, 17; 23; 31)$ | $b_x g_x b_x$ $(17; 24; 31)$ |

□

**Theorem A.8.** *There exists a plan that solves the planning problem $P_7(F)$ if and only if there exists an assignment $\sigma$ that satisfies the CNF formula $F$.*

*Proof.* $\Leftarrow$: Given an assignment $\sigma$ that satisfies $F$, construct an admissible plan whose induced formula assignment $\sigma_\Pi$ equals $\sigma$, by choosing the sequence of values of $v_s$ accordingly. It follows that for each clause $C_i$, $T_i \leq n$, since there exists a variable $x_j$ such that $\sigma_\Pi(x_j) = m_j$ satisfies $C_i$. Since $n \geq T_i$, there exists an admissible plan $\Pi$ for which $Q_i^{2n+1}$ has the form indicated in Case (b) or (c) of Lemma A.7. In either case, the $(2n+1)$-th value of variable $v_{in}^2$ is $g_x$, as required by the goal state. The plan $\Pi$ thus solves $P_7(F)$.

$\Rightarrow$: Let $\Pi$ be a plan that solves the planning problem $P_7(F)$. By Lemma A.5 the plan $\Pi$ is admissible. We show by contradiction that $\sigma = \sigma_\Pi$ satisfies $F$. Assume not. Then there exists a clause $C_i$ not satisfied by $\sigma$. Thus, Lemma A.7 applies to the sequence of values $Q_i^{2n+1}$ of $\Pi$. In particular, this means that the value of $v_{in}^2$ following the execution of $\Pi$ is $b_x$ according to Case (a) of the lemma. This contradicts $\Pi$ solving $P_7(F)$, since $b_x$ is different from the goal state $\mathsf{goal}(v_{in}^2) = g_x$. □

**Proposition A.9.** *Plan existence for $\mathbb{C}_n^7$ is NP-hard.*

*Proof.* The largest variable domains of the planning problem $P_7(F)$ are those of variables $v_{11}^1, \ldots, v_{kn}^3$, which contain 7 values. The proof follows immediately from the NP-hardness of CNF-SAT, Theorem A.8, and the fact that we can produce the planning problem $P_7(F)$ in polynomial time given a CNF formula $F$. $\qquad\square$

# References

Brafman, R., & Domshlak, C. (2003). Structure and Complexity in Planning with Unary Operators. *Journal of Artificial Intelligence Research, 18*, 315–349.

Brafman, R., & Domshlak, C. (2006). Factored Planning: How, When, and When Not. In *Proceedings of the 21st National Conference on Artificial Intelligence*, pp. 809–814.

Bylander, T. (1994). The computational complexity of propositional STRIPS planning. *Artificial Intelligence, 69*, 165–204.

Chapman, D. (1987). Planning for conjunctive goals. *Artificial Intelligence, 32(3)*, 333–377.

Chen, H., & Giménez, O. (2008). Causal Graphs and Structurally Restricted Planning. In *Proceedings of the 18th International Conference on Automated Planning and Scheduling*, pp. 36–43.

Domshlak, C., & Dinitz, Y. (2001). Multi-Agent Off-line Coordination: Structure and Complexity. In *Proceedings of the 6th European Conference on Planning*, pp. 277–288.

Erol, K., Nau, D., & Subrahmanian, V. (1995). Complexity, Decidability and Undecidability Results for Domain-Independent Planning. *Artificial Intelligence, 76(1-2)*, 75–88.

Giménez, O., & Jonsson, A. (2008a). The Complexity of Planning Problems with Simple Causal Graphs. *Journal of Artificial Intelligence Research, 31*, 319–351.

Giménez, O., & Jonsson, A. (2008b). In Search of the Tractability Boundary of Planning Problems. In *Proceedings of the 18th International Conference on Automated Planning and Scheduling*, pp. 99–106.

Helmert, M. (2006). The Fast Downward Planning System. *Journal of Artificial Intelligence Research, 26*, 191–246.

Jonsson, A. (2007). The Role of Macros in Tractable Planning Over Causal Graphs. In *Proceedings of the 20th International Joint Conference on Artificial Intelligence*, pp. 1936–1941.

Jonsson, P., & Bäckström, C. (1998). Tractable plan existence does not imply tractable plan generation. *Annals of Mathematics and Artificial Intelligence, 22(3–4)*, 281–296.

Katz, M., & Domshlak, C. (2008a). New Islands of Tractability of Cost-Optimal Planning. *Journal of Artificial Intelligence Research, 32*, 203–288.

Katz, M., & Domshlak, C. (2008b). Structural Patterns Heuristics via Fork Decompositions. In *Proceedings of the 18th International Conference on Automated Planning and Scheduling*, pp. 182–189.

Knoblock, C. (1994). Automatically generating abstractions for planning. *Artificial Intelligence, 68(2)*, 243–302.

Williams, B., & Nayak, P. (1997). A reactive planner for a model-based executive. In *Proceedings of the 15th International Joint Conference on Artificial Intelligence*, pp. 1178–1185.